# The Evolution of Sentiment Analysis - A Review of Research Topics, Venues, and Top Cited Papers

Mika V. Mäntylä[1] Daniel Graziotin[2], Miikka Kuutila [1]
[1] M3S, ITEE, University of Oulu
mika.mantyla@oulu.fi,
[2] Institute of Software Technology, University of Stuttgart

Corresponding author:
Mika Mäntylä,
Telephone: +358505771684
mika.mantyla@oulu.fi
P.O. Box 9210
FI-90014
Finland

## Abstract

Sentiment analysis is one of the fastest growing research areas in computer science, making it challenging to keep track of all the activities in the area. We present a computer-assisted literature review, where we utilize both text mining and qualitative coding, and analyze 6,996 papers from Scopus. We find that the roots of sentiment analysis are in the studies on public opinion analysis at the beginning of 20th century and in the text subjectivity analysis performed by the computational linguistics community in 1990's. However, the outbreak of computer-based sentiment analysis only occurred with the availability of subjective texts on the Web. Consequently, 99% of the papers have been published after 2004. Sentiment analysis papers are scattered to multiple publication venues, and the combined number of papers in the top-15 venues only represent ca. 30% of the papers in total. We present the top-20 cited papers from Google Scholar and Scopus and a taxonomy of research topics. In recent years, sentiment analysis has shifted from analyzing online product reviews to social media texts from Twitter and Facebook. Many topics beyond product reviews like stock markets, elections, disasters, medicine, software engineering and cyberbullying extend the utilization of sentiment analysis[1].

## 1 Introduction

*"The pen is mightier than the sword"* proposes that free communication (particularly written language) is a more effective tool than direct violence [1]. Sentiment analysis is a series of methods, techniques, and tools about detecting and extracting subjective information, such as opinion and attitudes, from language [2]. Traditionally, sentiment analysis has been about opinion polarity, i.e., whether someone has positive, neutral, or negative opinion towards something [3]. The object of sentiment analysis has typically been a product or a service whose review has been made public on the Internet. This might explain why sentiment analysis and opinion mining are often used as synonyms, although, we think it is more accurate to view sentiments as emotionally loaded opinions.

The interest on other's opinion is probably almost as old as verbal communication itself. Historically, leaders have been intrigued with the opinions of their subordinates to either prepare for opposition or to increase their popularity. Examples of trying to detect internal dissent can be found already at Ancient Greece's times [4]. Ancient works in East and West mingle with these subjects. "The Art of War" has a chapter on espionage that deals with spy recruiting and betrayal, while in the beginning of "Iliad" the leader of Greeks Agamemnon tries to gauge the fighting spirit of his men. Voting as a method to measure public opinion on policy has its roots in the city state of Athens in the 5$^{th}$ century BCE [5]. Efforts in capturing public opinion by quantifying and measuring it from questionnaires have appeared in the first decades of twentieth century [6], while a scientific journal on public opinion was established in 1937 [7].

We have seen a massive increase in the number of papers focusing on sentiment analysis and opinion mining during the recent years. According to our data, nearly 7,000 papers of this topic have been published and, more interestingly, 99% of the papers have appeared after 2004 making sentiment analysis one of the fastest growing research areas. Although the present paper focuses on the research articles of sentiment analysis, we can see that the topic is getting attention in the general public, as well. Figure 1 shows the increase in searches made with a search string “sentiment analysis” in Google search engine.

We observed that the first academic studies measuring public opinions are during and after WWII and their motivation is highly political in nature [8], [9]. The outbreak of modern sentiment analysis happened only in mid-2000’s, and it focused on the product reviews available on the Web, e.g., [3]. Since then, the use of sentiment analysis has reached numerous other areas such as the prediction of financial markets [10] and reactions to terrorist attacks [11]. Additionally, research overlapping sentiment analysis and natural language processing has addressed many problems that contribute to the applicability of sentiment analysis such as irony detection [12] and multi-lingual support [13]. Furthermore, with respect to emotions[2], efforts are advancing from simple polarity detection to more complex nuances of emotions and differentiating negative emotions such as anger and grief [15].

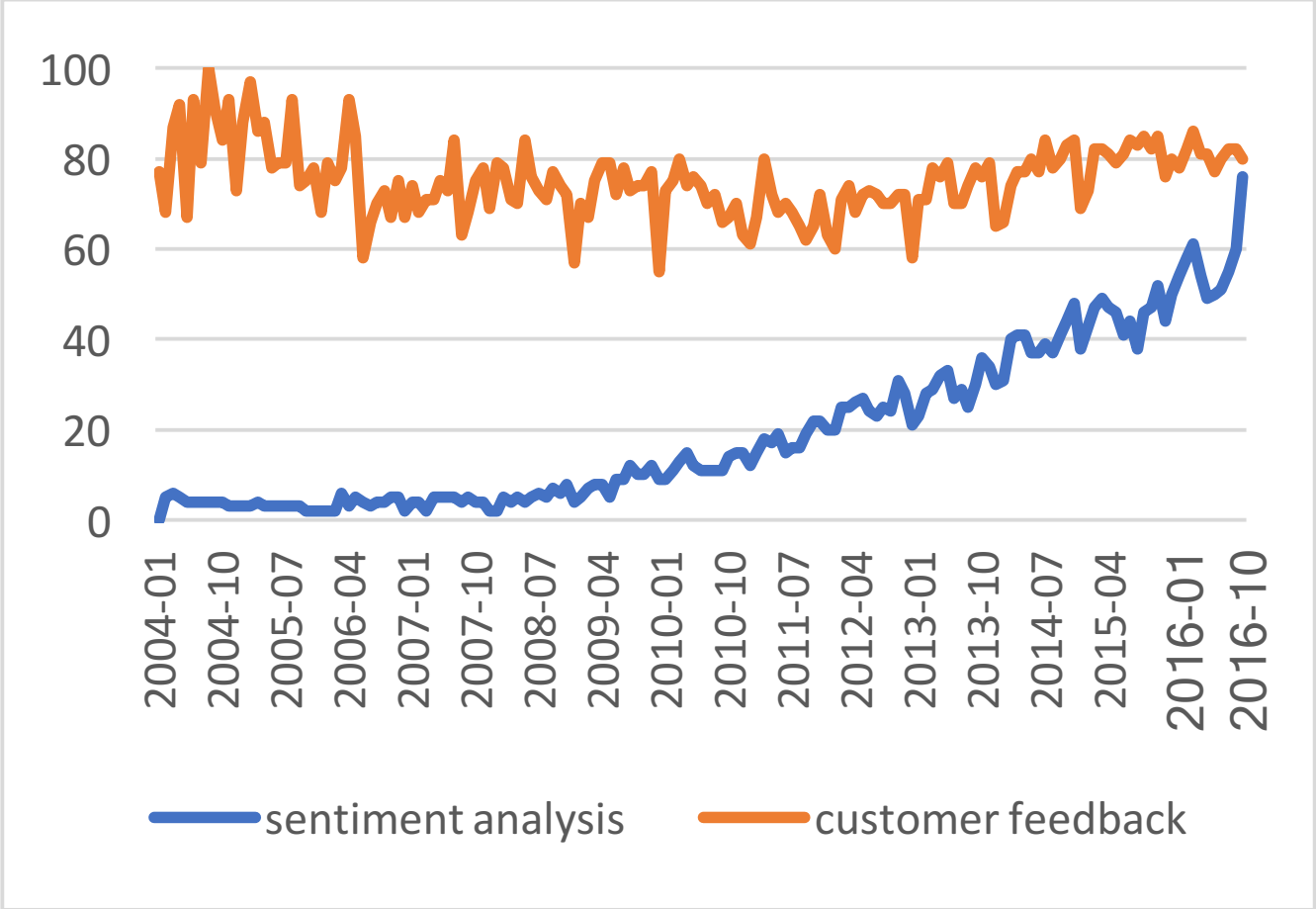

**Figure 1 Google Trends (www.google.com/trends) data showing the relative popularity of search strings “sentiment analysis” and “customer feedback”.**

The area of sentiment analysis has become so large that any individual researcher would face several issues when keeping track of all the activities in the area and the information overload. An academic literature review can only focus on one particular area of sentiment analysis as it typically includes between 10 to 100 studies, e.g., a recent systematic review of the prediction of financial markets with sentiment analysis reviewed 24 papers [10]. To overcome the challenges caused by the increasing number of articles about sentiment analysis, we present a computer-assisted literature review and a bibliometric study of sentiment analysis. Studies like the one we are presenting should be helpful when working in an area with large volumes of literature.

We think that the present article can offer an overview of sentiment analysis to newcomers and it may provide valuable to more seasoned scholars for educational purposes. To provide such an overview, we characterize the field of sentiment analysis by answering the following research questions that are typical in bibliometric studies, e.g., [16]–[19]:

- *RQ1: What is the number of papers in sentiment analysis?* This question helps us in understanding the volume of work in sentiment analysis. We can also observe yearly trends that tell us about the history of the topic and can help in predicting the future.
- *RQ2: What is the number of citations in sentiment analysis?* This question addresses the impact of sentiment analysis. Similarly to RQ1, we can observe yearly trends about the history of the topic and can also help in predicting the future.

[2] See [14] for the differences between feelings, opinions, sentiment, and emotions in the context of natural language processing.

- *RQ3: What are the most popular publication venue for sentiment analysis?* This question shows the popular venues for publishing sentiment analysis studies. Understanding the different communities related to sentiment analysis helps understanding the entire field.
- *RQ3: What research topics have been investigated with sentiment analysis?* Given that the topic has rapidly grown very large, we use text clustering to get on overview of the different areas of sentiment analysis. Our text clustering approach originates from influential paper by Griffiths and Steyvers [20] titled "Finding scientific topics". We support the automated text clustering with manual qualitative analysis and they jointly provide a thematic overview of the research topics in this field.
- *RQ4: What are the most cited original works and literature reviews in sentiment analysis and what research topics do these papers cover?* Citations are often referred as the backbone of science. Investigation of the landmarks in sentiment analysis can demonstrate the most interesting and impactful work in this area. Using citation counts has been previously done for example [21], [22] that studied the 100 most cited papers published in Nature and in Software Engineering.

The paper makes the following six contributions. First, we show how attempts to understand public opinion at the start of $20^{th}$ century through questionnaires and subjectivity analysis in computational linguistics gave birth to this topic. For decades, the research area was mostly ignored until massive amounts of opinions available in the Web gave birth to modern sentiment analysis. Second, we demonstrate how modern sentiment analysis has received a 50-fold growth in ten years between 2005 and 2016. The number of citations has grown along with the paper counts. Third, we show that most popular venues for sentiment analysis are series with large volumes: Lecture Notes In Computer Science, CEUR Workshop, ACM International Conference Proceeding Series. However, the venues with the highest shares of sentiment analysis papers adjusted by the total publication output are Procesamiento De Lenguaje Natural, International Conference Recent Advances In Natural Language Processing, and Communications In Computer And Information Science. Newcomers have now an indication of venues that welcome the topic. Fourth, using word clouds we show that most prominent research activities have been related to the classification of online reviews and social media texts. We demonstrate using how the research activities have shifted from analyzing online product reviews, which were more common prior to 2014, to social media texts that dominate the studies in the years 2014-2016. Fifth, we demonstrate the value of mixing Latent Dirichlet Allocation (LDA, see section 2.4) topic modelling with qualitative coding for clustering all sentiment analysis papers and constructing a comprehensive classification of articles. This technique allows a deeper understanding of different research topics of an area, sentiment analysis in our case, by providing a tree like semantic structure. Sixth, we review the top-cited papers according Scopus and Google Scholar to show the hallmarks of sentiment analysis research.

This paper is structured as follows. Section 2 explains our research methods. Beyond that our structure deviates from conventional research articles. The related work in the area of sentiment analysis is an output of the results in Section 3. Section 4 discusses the limitation of this work and Section 5 provides the conclusions.

# 2 Research Methods

In this section, we present the research methods that we used in order to answer our research questions. Section 2.1 explains our search strategy. Section 2.2 describes the quantitative analysis while Section 2.3 tells how we analyzed the publication venues. In Section 2.4, we explain how we studied the research topics with automated text analysis and Section 2.5 explains the manual qualitative classification we did on top of the automated analysis. Finally, Section 2.6 describes our analysis of the top-cited papers.

## 2.1 Searching literature

We used Scopus search engine for our literature search and data retrieval. Our results in Sections 3.1-3.6 are based on Scopus data. According to its developer Elsevier, "Scopus is the largest abstract and citation database of peer-reviewed literature: scientific journals, books and conference proceedings" [3]. Thus, it should offer the widest coverage of scientific literature that one can achieve with a single search engine. We are aware that there has been much discussion about which academic search database offers the least error-prone results and the highest coverage for knowledge discovery and researchers' assessment, e.g., [23]–[26]. Later in Section 4, we report how Scopus was found to be the most reliable [26] and comprehensive [25] academic search engine that is compatible with our aims. Scopus also offers advanced search engine features such as finding variant spellings. Another benefit of using Scopus is that it allows downloading paper titles, and abstracts in batches of 2,000 papers at a time. This enables further offline analysis, for example in the form of text clustering. The batch export functionalities are limited in Google Scholar and Web of Science which thus become less usable for our purposes.

[3] https://www.elsevier.com/solutions/scopus

We used several search strings to search the literature in Scopus, see Table 1. Two of our search string originated from the highly influential literature review on sentiment analysis by Pang and Lee [27] that presented strings "Opinion mining" and "Sentiment analysis" as synonyms or near synonyms. Additionally, we searched the literature about historical terms, e.g. semantic orientation. All of our search strings were targeted to the paper title, paper abstract, and paper keywords and combined with OR operator in Scopus. A match any of our search string in either title, abstract or key words resulted in the inclusion of a paper.

We used Google Scholar in addition to Scopus to search for the top-cited papers as the papers from the early years of sentiment analysis are not indexed by Scopus. Our results in two sections, Section 3.1 and 3.6, use data from both Google Scholar and Scopus. In more detail, we executed 11 queries corresponding to the search strings in Table 1 with Publish or Perish 5 software. We retrieved the top 100 results sorted by relevance for each string as Google Scholar does not allow sorting based on citations. As Google Scholar's relevance also considers citations, we are confident that we were able to estimate the top cited sentiment analysis papers from although there is no way to guarantee this. These 1,100 hits (11*100) were exported from Publish or Perish software to a CSV-file and merged and sorted for citations. This additional check should ensure that historical hallmark papers of sentiment analysis have not been missed when reconstructing the history of sentiment analysis.

**Table 1 Search strings and the share of search hits in Scopus (with overlaps)**

| Search term | % of hits |
|---|---|
| "sentiment analysis" | 68.5 % |
| "opinion mining" | 29.1 % |
| "sentiment classification" | 18.0 % |
| "opinion analysis" | 5.6 % |
| "semantic orientation" | 3.8 % |
| sentiwordnet | 2.7 % |
| "opinion classification" | 1.4 % |
| "sentiment mining" | 1.3 % |
| "subjectivity analysis" | 1.1 % |
| sentic | 1.0 % |
| "subjectivity classification" | 0.8 % |

## 2.2 Quantitative analysis of paper and citation counts

We performed quantitative analysis by plotting histograms of the paper and citation counts. Our analysis scripts are openly available[4].

## 2.3 Analysis of publication venues

Given the large body of work we were interested in finding out the publication venues of the papers in order to further demarcate the publication area. Computer science disciplines publish the majority of research results into conference proceedings instead of journals[5]. While journals rarely change their title name and vary in issue and volume numbers, we discovered that conferences proceedings names are not reliable over the years. In order to overcome this issue, the second author cleaned the venues that we retrieved from Scopus using R language [29]. The cleanup criteria included the deletion of the years from the conference names as well as their enumeration (e.g., 2015, 1st, 22nd, etc.). We also removed substrings related to the term *proceedings* (e.g., "Proc. of the ", "Proceedings of the "), because the term was also not used consistently over the years by the same conferences. After the cleanup, we found that overall the papers were distributed across 1,526 different sources.

4 https://doi.org/10.6084/m9.figshare.5537176

5 Articles published in journals are lately becoming as common as those in conferences, although it is still not the case yet [28].

## 2.4 Word clouds and Topic modelling with LDA of research topics

Due to our high volume of articles (nearly 7,000) qualitative manual analysis of all the papers would have exceeded our resources. Therefore, we used the text mining and clustering with R language [29]. We did basic word clouds and dissimilarity word clouds using R package “wordcloud” [30].

We used Latent Dirichlet Allocation (LDA) topic modelling to cluster our papers. LDA is a soft clustering algorithm developed for several scopes including text clustering. LDA approaches text clustering by acknowledging that each text document can be about multiple topics, e.g., a document describing cat foods would be about two more general topics, namely cats and foods. LDA tries to model how or from what topics a particular text corpus could have been created from. In our case, we could for example find that sentiment analysis corpus is created from topics such as the application area, e.g., hotel reviews, the machine-learning algorithm used, the natural language processing techniques used and so on. Thus, as opposed to hard clustering where each paper would be assigned to a single topic only, LDA soft clustering gives shares (expressed by the Greek letter θ (theta) in the formal description of LDA algorithm, see [20]) that expresses how many of the words of a particular document come from a certain topic. A topic in LDA is nothing more than a collection of words and their probability estimates. For example, a paper analyzing hotel reviews with a Bayesian algorithm could have the highest shares of words coming from two topics one containing words “hotel”, “reviews”, and the second containing words “Bayesian”, “algorithm”.

In more detail, we used LDA from R package “topicmodels” [31]. The origins of the LDA approach for scientific topic detection lie in the influential paper by Griffiths and Steyvers [20] modelling scientific topics that was published in PNAS. Parts of the R computations are from Ponweiser [32] who replicated the study by Griffiths and Steyvers. Previously, we followed this approach when we analyzed software engineering literature [16].

We followed the approaches and advices of the papers in the previous section and proceeded as follows. First, we removed all publisher’s copyright information in the abstract. This text occasionally contains publisher names, e.g., IEEE or ACM. Second, we created and manipulated our corpus with R-package “tm” [33]. The corpus was created by merging the title and the abstract of each article. Then, we performed the following preprocessing steps: removed punctuation and numbers, made all the letters lower case, removed common stop words for English, and finally created a document-term-matrix aka a document-word-matrix, which describes the frequency of terms, while removing words with less than three letters and words occurring less than 5 times as in [32]. With the document-term-matrix, we performed term-frequency inverse-document-frequency (tf-idf) computations. Tf-idf improves raw term frequency computations by also considering the idf of each term that can be characterized as the amount of information each word carries. We used this information to remove the words having less than median tf-idf value from the document-term-matrix. An example of this step is show in [31]. In our experience, such tf-idf pre-processing reduces the computing time needed for creating LDA topics while improving the coherence the topics as less relevant words are not considered in the clustering. We used hyper-parameter settings with the default values suggested by Griffiths and Steyvers [20], i.e., alpha=50/k and beta=0.1. The hyper-parameters describe the Bayesian prior beliefs about how likely is each document to contain multiple topics (alpha) and how likely is each topic to contain to contain multiple words (beta). As in [20] we have fixed small beta that should lead to a quite high number of fine grained topics. Such specific topics were then furthered analyzed with qualitative coding, see next section. Finally, similar to [20] we tuned for k, i.e., finding the optimal number of topics (k) using a log-likelihood measure.

## 2.5 Qualitative coding of automatically created research topics

Since our LDA topic modelling from the previous step suggested that 108 topics would be optimal, we opted to create 108 topics. As understanding 108 topics as a flat list is difficult, we used qualitative coding strategy to improve readability and understandability. That is, we categorized the topics using qualitative coding as presented in many qualitative analysis handbooks, e.g., [34]. Our qualitative analyses were aided further by the investigation of the papers assigned to cluster and by the clustering of the LDA topics with hierarchical clustering of the topics and subsequent dendrogram visualizations. However, dendrograms were used as input only for the first author who performed the majority of the qualitative coding. First author was aided by the second author who did part of the qualitative coding and validated the qualitative coding performed by the first author. Finally, the first and second author jointly developed the qualitative coding strategy. During the qualitative coding phase the mind mapping tool *Mindmup.com* was used to enable real-time collaboration between the first and second author. Figure 2 shows a snap-shot of the *Mindmup.com* tool as it was used during qualitative coding. One can also see some of the topics produced by the LDA topic modelling, i.e., bubbles starting with the text “Topic”, and the emerging qualitative classification.

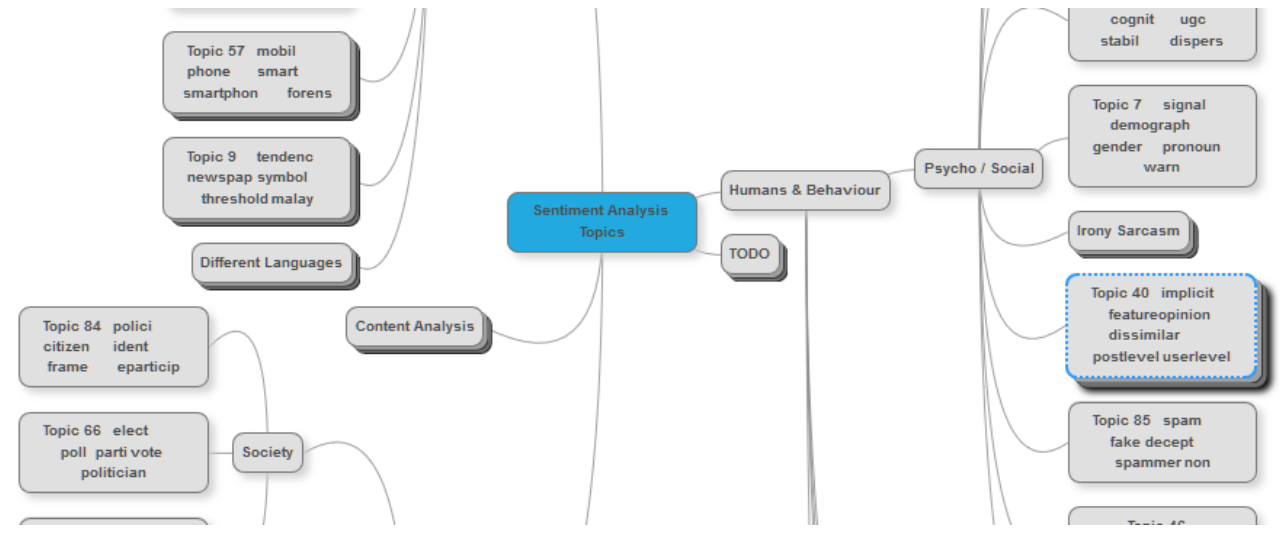

**Figure 2 Mindmup.com tool was used to collaborate in quality coding of the topics.**

### 2.6 Analysis of the top-cited papers

In the final phase, the third author studied and summarized the top-cited papers in our data set. The classification created in the previous phase was utilized to show which parts of the classification had been studied in the most cited papers. To find out the top cited papers we used citation counts normalized for time from Scopus and Google Scholar.

## 3 Results

### 3.1 Number of papers and a brief history of sentiment analysis (RQ1)

The number of papers in sentiment analysis is increasing rapidly as can be observed from Figure 3. The field has also changed in terms of content over the years. Prior to availability of massive amount of text and opinions online, studies mainly relied on survey-based methods and were focused on public or expert opinions rather than users or customers' opinions. The first paper that matched our search was published in 1940, and it was titled "The Cross-Out Technique as a Method in Public Opinion Analysis" [8]. In 1945 and 1947 three papers appeared that addressed measuring public opinions in post WWII countries that had suffered during the war (Japan, Italy, and Czechoslovakia) and they were all published in the journal Public Opinion Quarterly [9], [35], [36].

In mid 90s, computer-based systems started to appear. The computing revolution started to be reflective in research, too. For example, a paper titled "Elicitation, Assessment, and Pooling of Expert Judgments Using Possibility Theory" was published in 1995, and it used a computer system for expert opinion analysis in the domain industrial safety that allowed for example a pooling of opinions [37]. Still, the outbreak of modern sentiment analysis was nearly ten years away.

Another branch of work that was highly influential to the birth of modern sentiment analysis was the work performed by the Association for Computational Linguistics founded in 1962. It was in that community that computer based sentiment analysis was mainly born in, and the earliest paper we could find was Wiebe that in 1990 proposed methods to detected subjective sentences from a narrative in 1990 [38] and later in 1999 proposed a gold standard to do this [39].

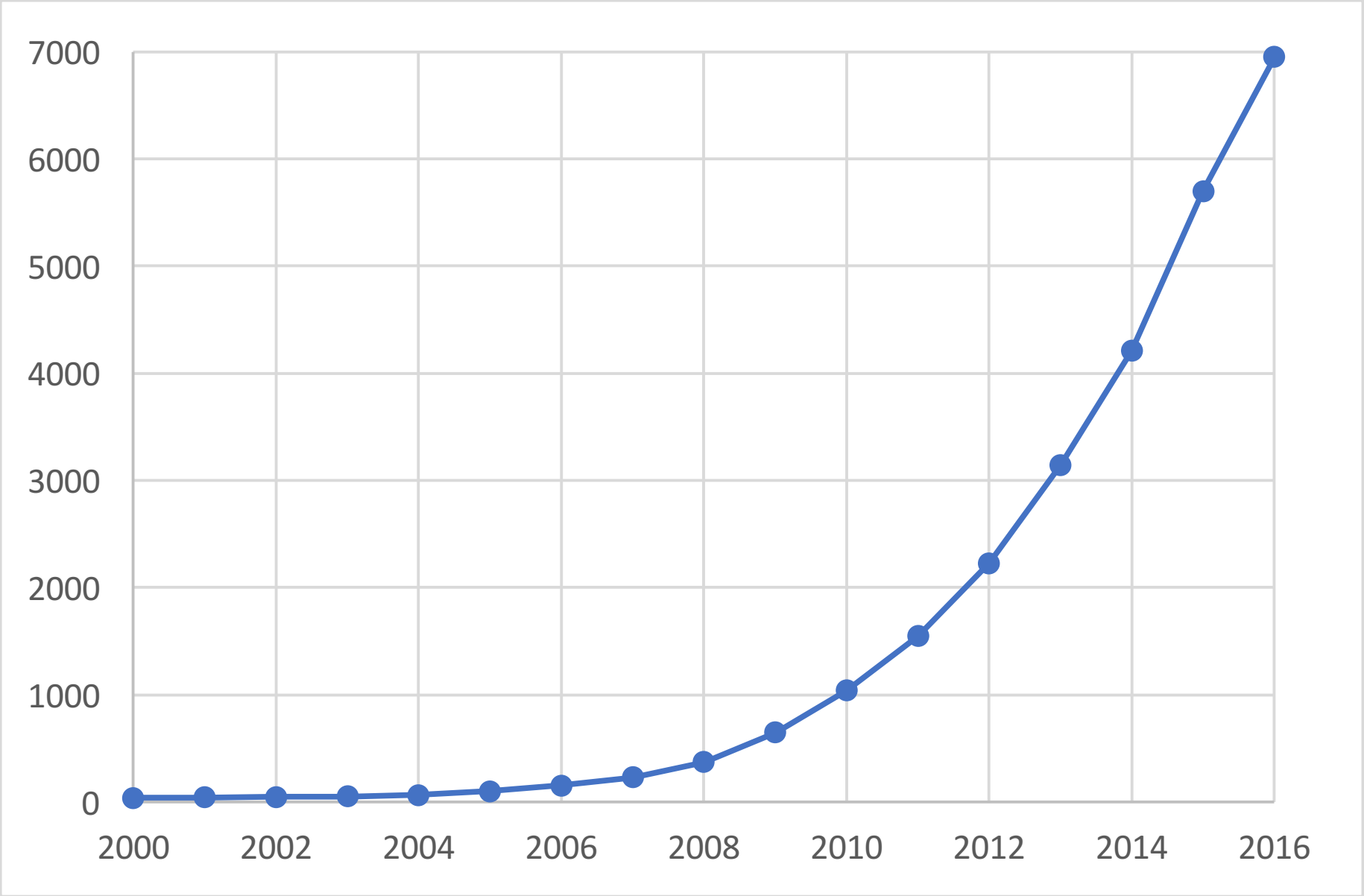

**Figure 3 Number of papers published per year in Scopus**

Being precise of what has been the first paper of modern sentiment analysis is hard as early years used fluctuating terminology. Using citation counts from Scopus and Google Scholar, we can pinpoint many influential papers from the turn of 21st century that started the modern sentiment analysis. In 1997 Hatzivassiloglou, and McKeown published a paper titled "Predicting the Semantic Orientation of Adjectives" where the authors used data from 1987 Wall Street Journal corpus and built a list of positive and negative adjectives and predicted whether conjoined adjectives are of the same or different orientation [40]. In 2002, using ratings from the web kick-started modern sentiment analysis. In 2002, a paper by Bang et al titled "Thumbs up? Sentiment Classification using Machine Learning Techniques" [41] used movie review data and found out that machine learning classification outperformed human-produced baselines. In the same year, Turney published a paper with a very similar title: "Thumbs Up or Thumbs Down? Semantic Orientation Applied to Unsupervised Classification of Reviews" [42] where he used online reviews of automobiles, banks, movies, and travel destinations and achieved on average accuracy of 74% for recommendations. In 2003, more influential papers were published, Turney and Litmann in a paper titled "Measuring Praise and Criticism: Inference of Semantic Orientation from Association" [43] proposed a method to automatically infer the semantical orientation of a word from the statistical context in which it appears in. Another influential paper from 2003 was titled "Mining the peanut gallery: Opinion extraction and semantic classification of product reviews" [3] that was published in the WWW conference. The paper positioned sentiment analysis in the context of product reviews available in the Web. The authors argued that automated analysis of such reviews is needed due to the high volumes of such reviews.

For years, the number of related published papers was extremely low. Figure 3 shows the published papers per year starting from the year 2000. By the year 2000 only 37 papers in total had been published, and by the year 2005 the number was 101. We can observe a momentum building up in 2005. By the end of 2010 the number of papers was 1,039 and, finally, by the end of 2015 the number of papers reached the value of 5,699. Our total pool of papers also includes many 2016 papers and the total count is at 6,996 papers. Our analysis shows that, although an understanding of opinions has always been important, what made opinion analysis a trending research topic was the possibility to automatically collect and analyze large corpuses of opinions with the help of text mining tools.

## 3.2 Publication Venues (RQ3)

The top 15 publication venues in terms of published sentiment analysis articles are in Table 2. The table provides the publication name, its type (P for proceedings, B for books, and J for journals), the number of published papers in 2015, the number of sentiment analysis articles and their percentage with respect to the entire sentiment analysis dataset, the ratio of the published sentiment analysis papers versus all published papers for the venue itself in 2015, and the number of citations received divided by the total sentiment analysis output of the venue itself . The fourth and fifth columns are bounded to the most recent year of complete data (2015) for publication because of Scopus limitations when retrieving data.

Despite of all our data cleanup efforts, we found out that Scopus enlists certain conference entries with the name of the proceedings series instead of the name of the conference itself. For example, the first four entries in Table 2 are proceedings series, which contain several different conferences, while the fifth entry represents the proceedings of a single conference. We first observe in Table 2 that most venues are conference proceedings (11 entries), while the remaining are journals (4 entries).

The most populated venue is Springer's Lecture Notes in Computer Science (LNCS) with its subseries Lecture Notes in Artificial Intelligence and Lecture Notes in Bioinformatics. LNCS publishes conference proceedings mainly, although the series disseminates nine transactions and other monographs such as Festschrifts, tutorials, and state of the art surveys. With 684 entries, LNCS has published almost four times sentiment analysis papers than the second most populated entry and it hosts 12.0% of all sentiment analysis papers indexed by Scopus. However, less than 1% of the 2015 LNCS papers were about sentiment analysis.

CEUR Workshop Proceedings (CEUR-WS) is a long-standing open access publication service in cooperation with Aachen University. CEUR-WS hosts exclusively proceedings of computer science workshops, which are free to publish and access for authors and readers, respectively. With 163 entries for CEUR-WS, we note that the two top venues (LNCS and CEUR-WS) account for half of the sentiment analysis papers of our top 15 rankings in Table 2. This indicates that the distribution of papers with respect to the venues is skewed towards the top two venues. However, the first two venues account for ca. 15% of all papers on sentiment analysis that are indexed by Scopus. Less than 1% of the 2015 CEUR-WS papers were about sentiment analysis.

The ACM International Conference Proceedings Series (ICPS) is an ACM program for cost-effective publication of conference and workshop proceedings hosted into ACM digital library. Conference organizers submit the proceedings to ACM, which evaluates them for inclusion into ICPS. We note that ACM main conferences, such as the International Conference on Software Engineering, the ACM CHI Conference on Human Factors in Computing Systems, and SIGGRAPH conference are not published with ICPS. The main conferences in computer science have their own proceedings, which are published by ACM, IEEE, or both. Less than 1% of the 2015 ICPS papers were about sentiment analysis.

Communications in Computer and Information Science (CCIS) is a Springer series of books devoted to the publication of conference proceedings of computer science conferences. As a way to differentiate from LNCS, CCIS declares to be interested mostly in topics related to theory of computing and information and communications science and technology. We observed that a difference in terms of topics does not seen to exist in the case of sentiment analysis. In 2015, 9.5% of the CCIS papers were sentiment analysis articles.

The International Conference on Information and Knowledge Management (CIKM) accepts papers about on information and knowledge management, as well as recent advances on data and knowledge bases. The conference is co-sponsored by ACM SIGWEB and SIGIR. While not explicitly calling for sentiment analysis related topics, the conference has recently had sessions dedicated to it. 1.6% of the 2015 papers in CIKM were about sentiment analysis.

The Conference on Empirical Methods in Natural Language Processing (EMNLP) is an international conference on natural language processing. EMNLP is organized by the SIGDAT special interest group for linguistic data, which is part of the Association for Computational Linguistics (ACL). The conference calls for papers related to SIGDAT community interests, and sentiment analysis is a solicited topic. Similarly to CCIS, ENMLP had a share of 7.3% published papers as sentiment analysis articles in 2015.

The Annual Meeting of the Association for Computational Linguistics is an ACL Conference, which calls for papers about intelligent systems and their interactions with humans using natural language, computational and linguistic properties of language, speech recognition and translation, and information retrieval and extraction. The conference calls for contributions on sentiment analysis and opinion mining.

Advances in Intelligent Systems and Computing (AISC), like LNCS and CCIS, is a recent series of Springer books that specialize in intelligent systems and intelligent computing. AISC contains mainly proceedings and few edited books.

Expert Systems with Applications is an Elsevier journal, which publishes papers related to expert and intelligent systems applied in industrial settings (including education and public sector). The journal deals with any kind of expert and intelligent systems applied in any industrial area. Data and text-mining is a solicited topic.

The International Conference on World Wide Web (WWW) provides a forum for discussion in regard to the standardization of World Wide Web associated technologies and their impact on society and culture. The conference calls for Web mining, human factors, social network analysis, and content analysis. Several sentiment analysis papers are presented at the conference each year. The WWW proceedings are published by ACM. In our ranking, Expert

Systems with Applications and the WWW conference published the same amount of sentiment analysis papers. Therefore, they share position number 9.

The International Journal of Applied Engineering Research publishes articles in all engineering areas. The journal is published by Research India Publication publisher. We notice a high number of articles published per year and that the journal charges both for publishing and accessing the articles, which could be read for free only for one year after publication. The journal and its publisher were reported to be of predatory publishing nature [44].

Procesamiento del Lenguaje Natural is an open access journal published by the Sociedad Española para el Procesamiento del Lenguaje Natural. The journal publishes articles on natural language processing. The journal welcomes articles in English as well in Spanish, although a bilingual abstract is required. The journal had the highest ratio of sentiment analysis papers in 2015 with a value of 28.6%.

The International Conference on Recent Advances in Natural Language Processing (RANLP) is an international conference organized by the Association for Computational Linguistics (ACL). Opinion mining and sentiment analysis are a welcomed topic for submission. 11.5% of the RANLP papers in 2015 was of sentiment analysis nature.

Decision Support Systems is an Elsevier journal. The journal welcomes submissions that are relevant to theoretical and technical issues in the support of decision support systems foundations, functionality, interfaces, implementation, and evaluation. While the journal does not call for data mining papers, papers related to sentiment analysis and opinion mining in the context of decision support systems are welcomed.

The International Conference on Data Mining (ICDM) is an IEEE conference which covers all aspects of data mining, including algorithms, software and systems, and applications. Opinion mining and sentiment analysis are naturally belonging to data mining, thus papers on those topics are solicited.

Should a potential author decide to submit to a venue for maximizing the number of citations instead, the top five venues would be the EMNLP Conference On Empirical Methods In Natural Language Processing, Expert Systems With Applications journal, the ACL Annual Meeting Of The Association For Computational Linguistics, Decision Support Systems journal, and the International Conference On World Wide Web. These venues are not the top 5 in terms of publication output. We advise potential authors to develop a publication strategy based on either (1) embracing of sentiment analysis, (2) potential research impact, or (3) both.

**Table 2 Publication Venues, number and shares of sentiment analysis papers and numbers of sentiment analysis papers in comparison to total number of papers published in 2015**

| Ranking | Name | Type | # of papers in 2015 | # of sentiment analysis papers | % of sentiment analysis papers in total | Proportional share of sentiment analysis papers (# of sentiment analysis papers in 2015 / # of papers in 2015) | Cites per sentiment analysis paper |
|---|---|---|---|---|---|---|---|
| 1 | Lecture Notes In Computer Science (including Subseries Lecture Notes In Artificial Intelligence And Lecture Notes In Bioinformatics) | P, B | 21,611 | 684 | 12.0% | 0.7% (145/21,611) | 1.65 |
| 2 | CEUR Workshop | P | 3,661 | 163 | 2.9% | 0.8% (29/3,661) | 0.47 |
| 3 | ACM International Conference Proceeding Series | P | 3,201 | 116 | 2.0% | 0.6% (19/3,201) | 1.13 |
| 4 | Communications In Computer And Information Science | P | 275 | 107 | 1.9% | 9.5% (26/275) | 1.09 |
| 5 | International Conference On Information And Knowledge Management | P | 246 | 90 | 1.6% | 1.6% (4/246) | 17.07 |
| 6 | EMNLP Conference On Empirical Methods In Natural Language Processing | P | 315 | 85 | 1.5% | 7.3% (23/315) | 33.87 |
| 7 | ACL Annual Meeting Of The Association For Computational | P | 356 | 66 | 1.2% | 3.1% (11/356) | 29.59 |
| | Linguistics | | | | | | |
| 8 | Advances In Intelligent Systems And Computing | P, B | 3,245 | 60 | 1.1% | 0.9% (30/3,245) | 0.45 |
| 9 | Expert Systems With Applications | J | 778 | 56 | 1.0% | 1.3% (10/778) | 30.43 |
| 9 | International Conference On World Wide Web | P | 134 | 56 | 1.0% | 1.5% (2/134) | 27.11 |
| 11 | International Journal Of Applied Engineering Research | J | 6,962 | 45 | 0.8% | 0.5% (37/6,962) | 0.13 |
| 12 | Procesamiento De Lenguaje Natural | J | 42 | 41 | 0.7% | 28.6% (12/42) | 2.10 |
| 13 | International Conference Recent Advances In Natural Language Processing | P | 96 | 38 | 0.7% | 11.5% (11/96) | 5.29 |
| 14 | Decision Support Systems | J | 130 | 36 | 0.6% | 2.3% (3/130) | 29.44 |
| 15 | IEEE International Conference On Data Mining | P | 146 | 33 | 0.6% | 2.0% (3/146) | 12.48 |

### 3.3 Citation patterns (RQ2)

As the number of papers grows so does the number of citations. We analyze the citation patterns as they are a backbone of science, which is well illustrated by the famous quote by Newton ("If I have seen further, it is by standing on the shoulders of giants"), meaning that only good prior works make research advances possible. We are also aware of the problems with citations counts and that citations are not always used as part of building on top of previous work [45]. Nevertheless, we consider them as useful for showing the growth in interest on sentiment analysis.

Figure 4 shows the cumulative total number of citations in comparison to the number of papers. The citations are counted for the papers that are published in the given year. Figure 5 shows the annual number of papers and the number of citations normalized for time. Since the accumulation of citations takes time, a normalization allows us to have more reliable data of the recent years. With respect to the year 2015 we assume that our time correction is insufficient in particular for papers that appeared late in 2015. Moreover, there is a delay between the publication date and the Scopus inclusion date. The fluctuation in the citation counts between years in Figure 5 is mainly due to single highly cited papers. For example, for the year 2008 one paper [27] has 2,487 citations out of the 4,771 total citations made for papers published on that year.

Figure 6 shows the number of citations given to each paper and the year when the paper was published. In the figure, we have used log scale transformation and we removed papers with zero citations to improve its readability. The figure also illustrates the recent growth this field has experienced in the past ten years.

Figure 7 shows a histogram of paper citations. We find that 54% of the papers (3,806/6,996) have not been cited at all, 13% papers (898/6,996) had one citation while 33% (2,292/6,996) had two or more citations. For comparison purposes, we report that our recent study of the software engineering literature showed that 43% of papers had no citations, 14% had one citations, and 43% had two or more citations [16]. In what follows, we compare the current results with relevant results of our previous study.

The difference in papers with no citations (sentiment analysis 57% vs. software engineering 43%) is due to the age of each research topic. The median publication year for papers in our previous software engineering data set was 2008 while in this paper the median is 2014. To adjust for age bias, we computed citation counts for papers aged between five and nine years for both data sets. This showed that in that age group software engineering has more papers with no citations (sentiment analysis 33% vs. software engineering 43%). Table 3 shows the percentage of not cited papers with respect to paper age in years.

Table 4 shows the top-5 cited papers from each area (here we use only Scopus papers as [16] included only Scopus data) with respective citations and publication years. A more detailed analysis showed how all of the top 30 cited papers in software engineering were published prior to 2005, while in the present study only five of the top 30 cited papers, were published prior to 2005. It is interesting to the present authors that the most cited paper in sentiment analysis has more citations than any paper in software engineering despite the only recent emergency of this research topic. This comparison further highlights the recency of sentiment analysis and opinion mining as a research topic.

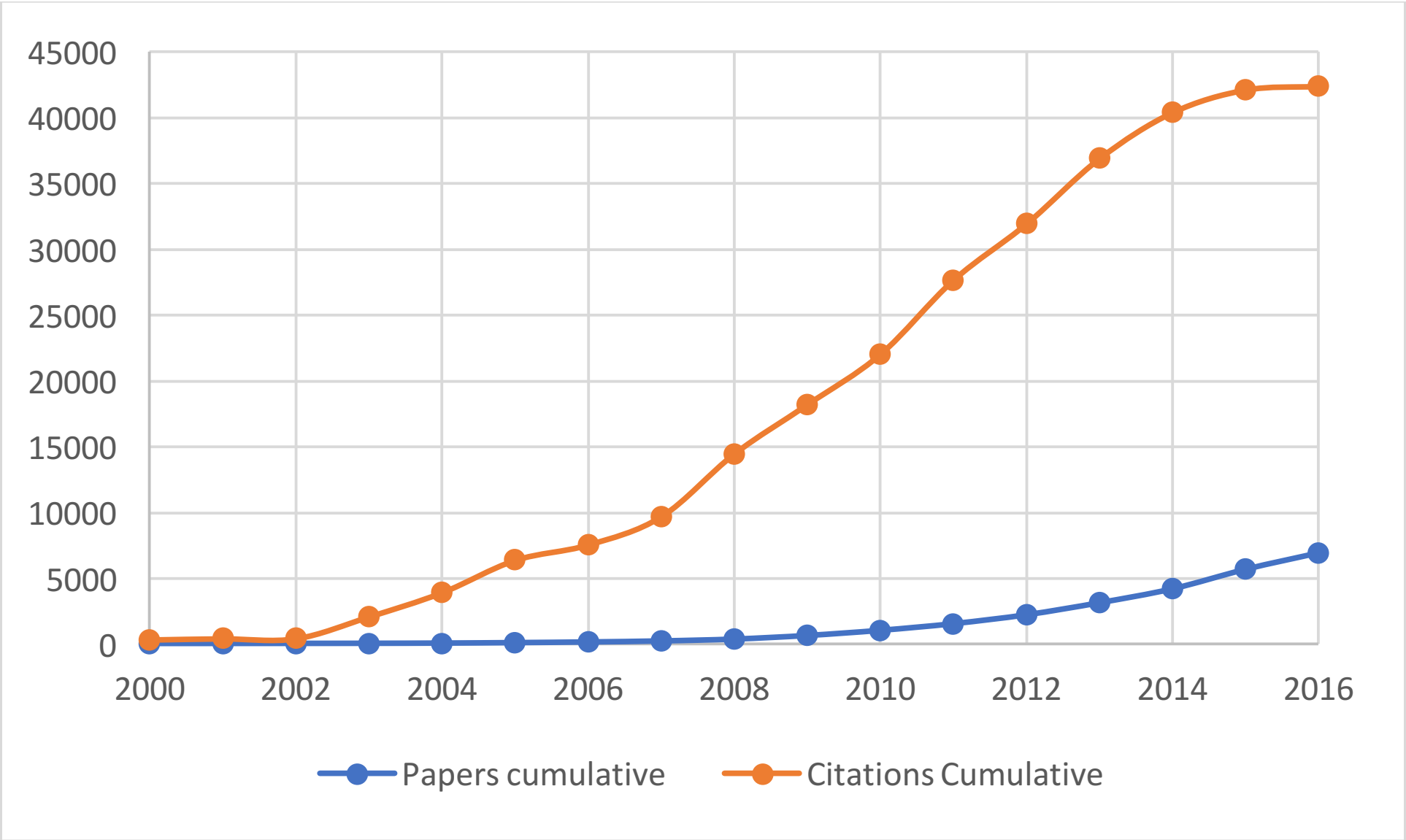

**Figure 4 Cumulative number of papers and citations**

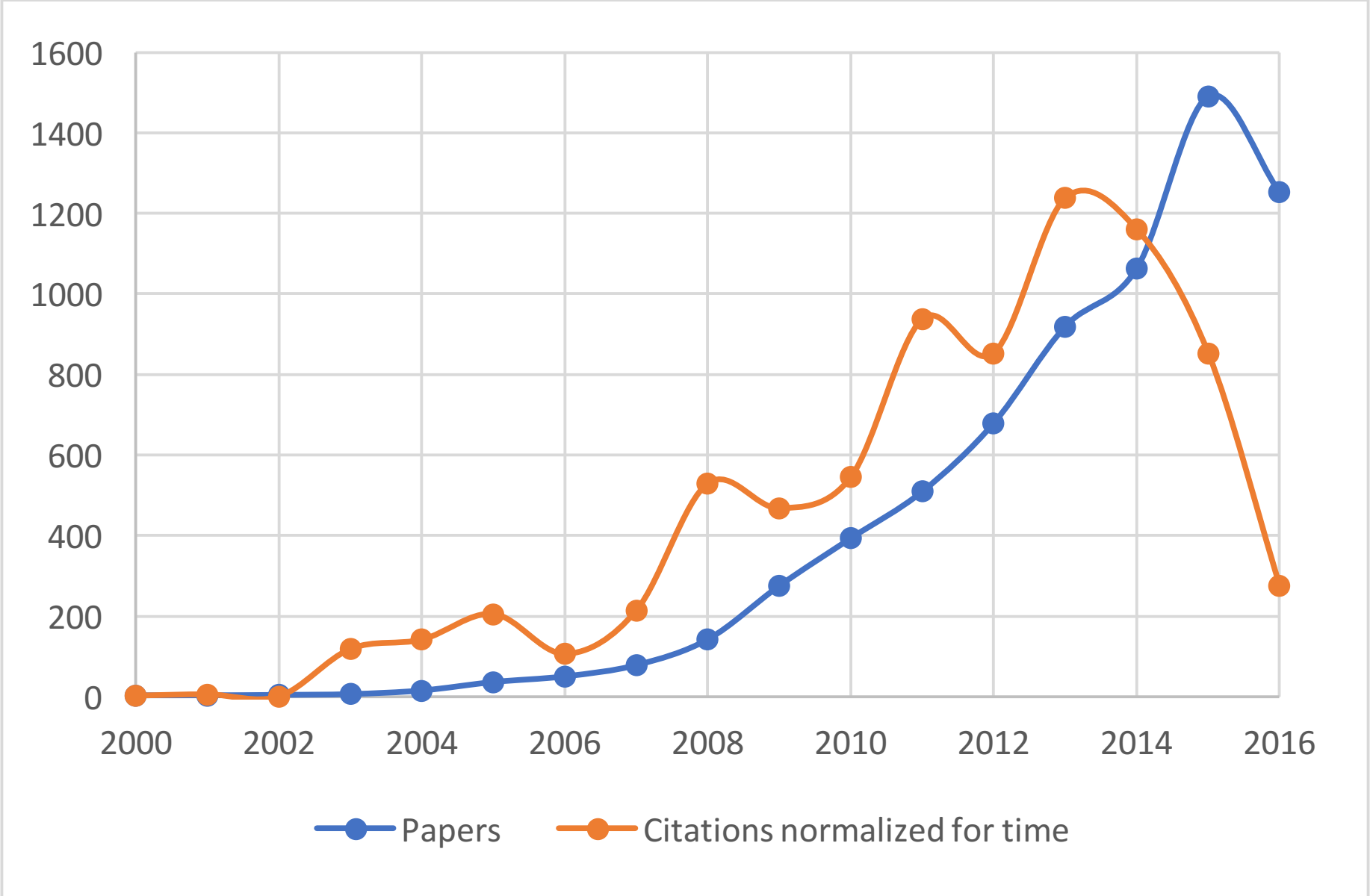

**Figure 5 Annual number of papers and citations**

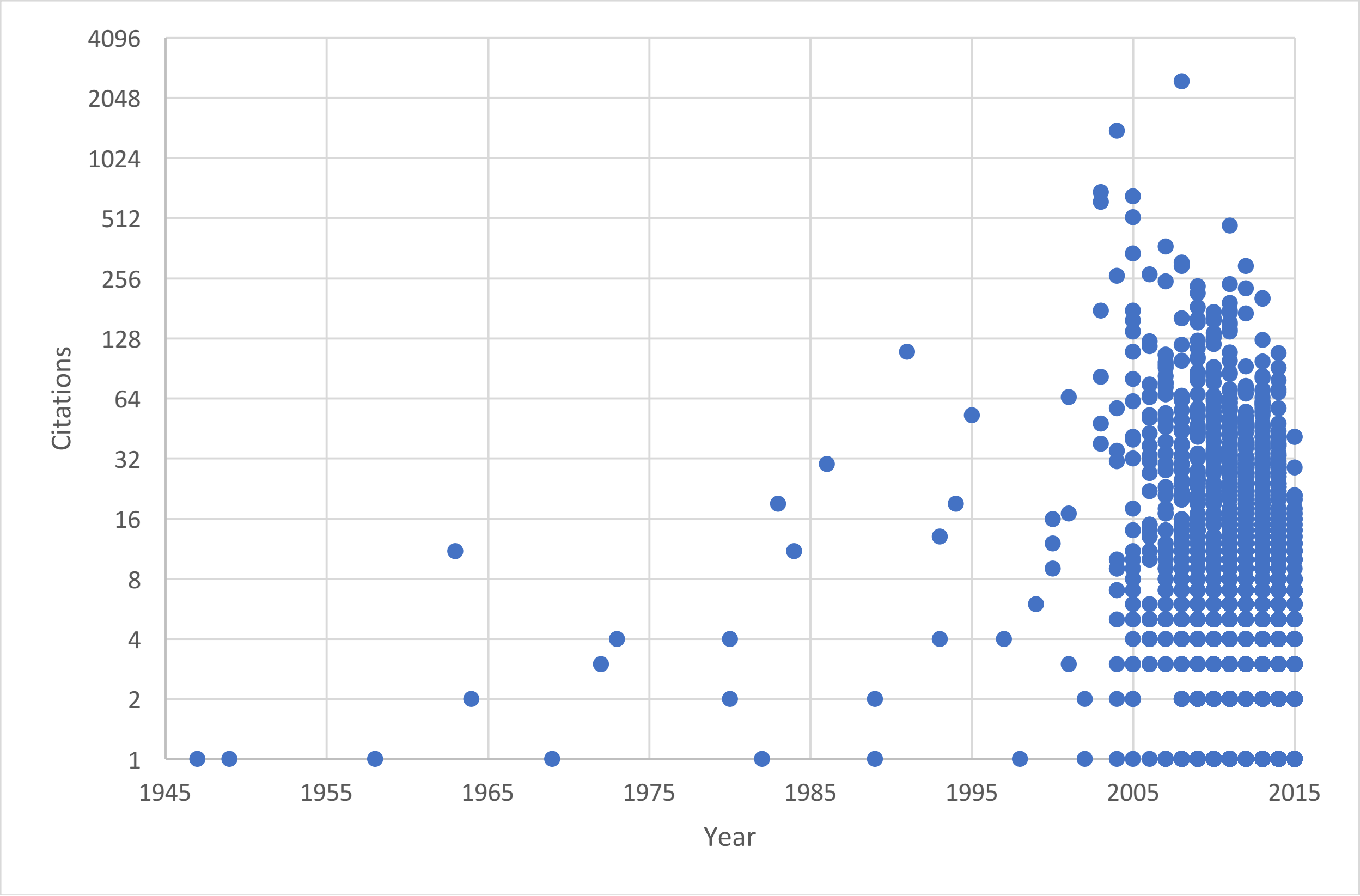

**Figure 6 Number of citations given to each paper and the year each paper was published**

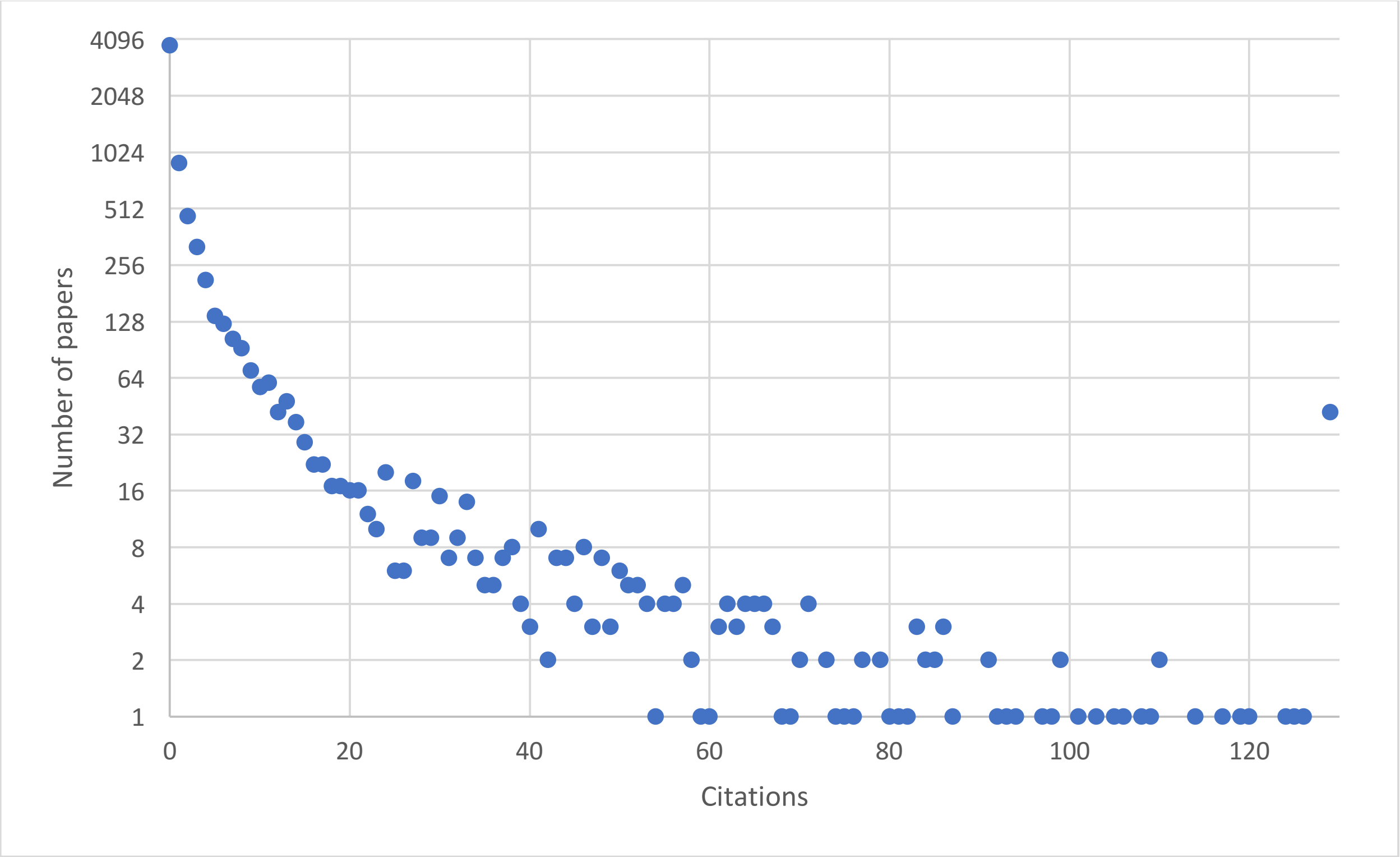

**Figure 7 Number of papers with given amount of citations (note the X-axis is not complete as the top-cited paper has been cited over 2000 times and it would make the entire figure unreadable)**

**Table 3 Percentage of not cited papers with respect to paper age in sentiment analysis and software engineering**

| Paper age (years) | 2 | 3 | 4 | 5 | 6 | 7 | 8 | 9 | 10 |
|---|---|---|---|---|---|---|---|---|---|
| Sentiment analysis | 51 % | 39 % | 41 % | 37 % | 33 % | 30 % | 34 % | 31 % | 34 % |
| Software engineering | 71% | 50% | 50% | 47% | 47% | 42% | 41% | 35% | 32% |

**Table 4 Top cited papers in Software Engineering [16] and in this study. Note: We only include papers found in Scopus here to make it comparable with previous work.**

| | Top-cited software engineering papers from [16] | | | Sentiment analysis and opinion mining | | |
|---|---|---|---|---|---|---|
| Rank | Title | Year | Citations | Title and citation | Year | Citations |
| 1 | A metrics suite for object-oriented design [46] | 1994 | 1,817 | Opinion mining and sentiment analysis [27] | 2008 | 2,487 |
| 2 | QoS - aware middleware for Web services composition [47] | 2004 | 1,696 | Mining and summarizing customer reviews [48] | 2004 | 1,400 |
| 3 | The model checker SPIN [49] | 1997 | 1,669 | Measuring praise and criticism: Inference of semantic orientation from association [43] | 2003 | 694 |
| 4 | Complexity measure [50] | 1976 | 1,304 | Recognizing contextual polarity in phrase-level sentiment analysis [51] | 2005 | 657 |
| 5 | Graph drawing by force directed placement [52] | 1991 | 1,162 | Mining the peanut gallery: Opinion extraction and semantic classification of product reviews [3] | 2003 | 617 |

### 3.4 Areas of research – Word clouds (RQ4)

Figure 8 shows the word cloud of all the paper titles. Figure 8 (a) shows how the key concepts of sentiment analysis are “social”, “online”, “reviews”, “media” and “product”. We can also find more detailed topics and domains such as “movie”, “news”, “political”, “stock” and “financial” that refer to the business domains. Also, different languages can be seen with “Chinese” and “Arabic” being the most common. Similarly, different analysis methods are visible in word like “neural”, “fuzzy, and “supervised”.

For an established research topic, one can show word clouds of different decades to highlight the differences. Due to the recency of our topic, we were not able to do that as the early years had very limited numbers of papers. Therefore, we divided the data to roughly two equal parts, that is the years 2014-2016 and the years 2013 and prior. Figure 8 (b) shows the differences between the more recent (2014-2016) and the early years (2013 and prior). Figure 8 (b) shows the word comparison cloud (the words on top in green are more common in 2014-2016 while the words on the bottom in orange are more common prior to 2014). Both figures are plotted with R Package ‘wordcloud’. In the word comparison cloud, the size of each word is determined by its deviation from the overall occurrence average. For example, word “reviews” becomes smaller meaning the difference is smaller between the periods. We can observe that sentiment analysis of “Chinese” language was more common in the early years while recent research activities highlight “Arabic” and “Indian” languages. Early years focused on “customer” “online” “product” reviews. While later years seem to focus on reviews made in “social” “media” and particularly in “Twitter” and “Facebook” (top left in Figure 8b). We can also observe that early years had focus on “web” and “online” while later years focus more on “mobile”. There has also been a shift in who produces the reviews. Early years referred the producers as “customer” while later years highlight the word “users”. We can see the word “big” coming from the big data concept. We note here that we have removed common words like results, data, sentiment, analysis, mining, opinion etc. to make the word clouds more meaningful.

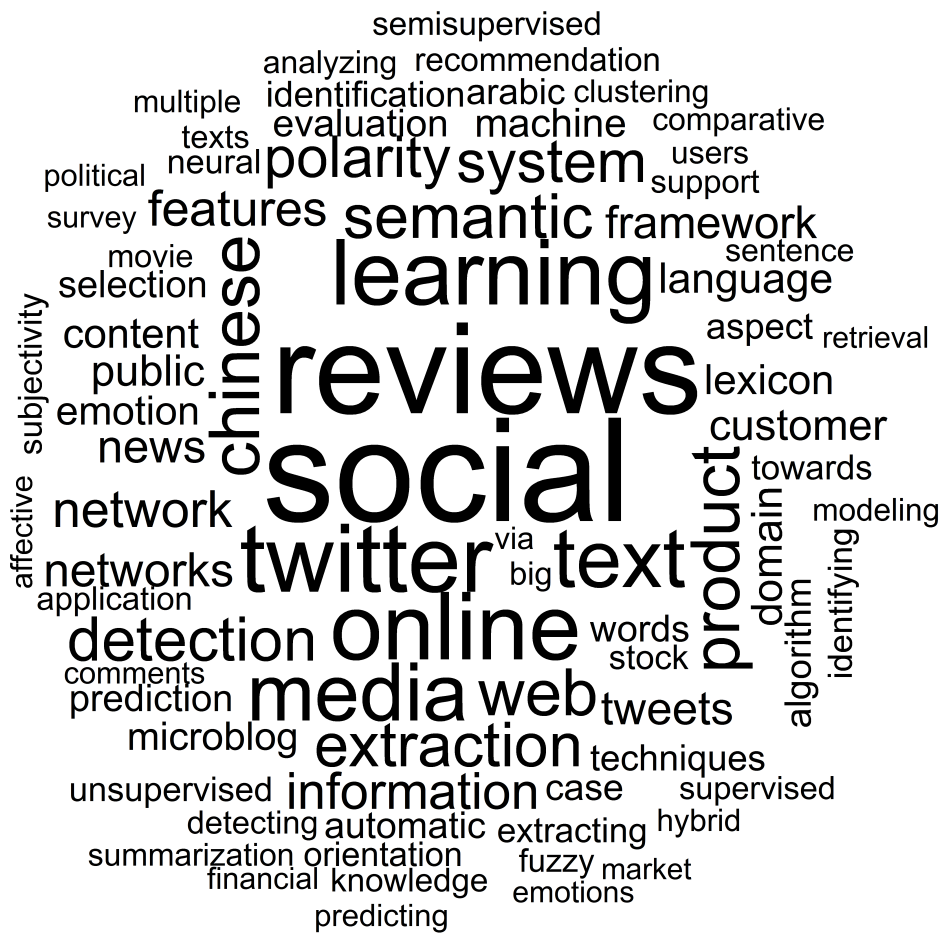

(a)

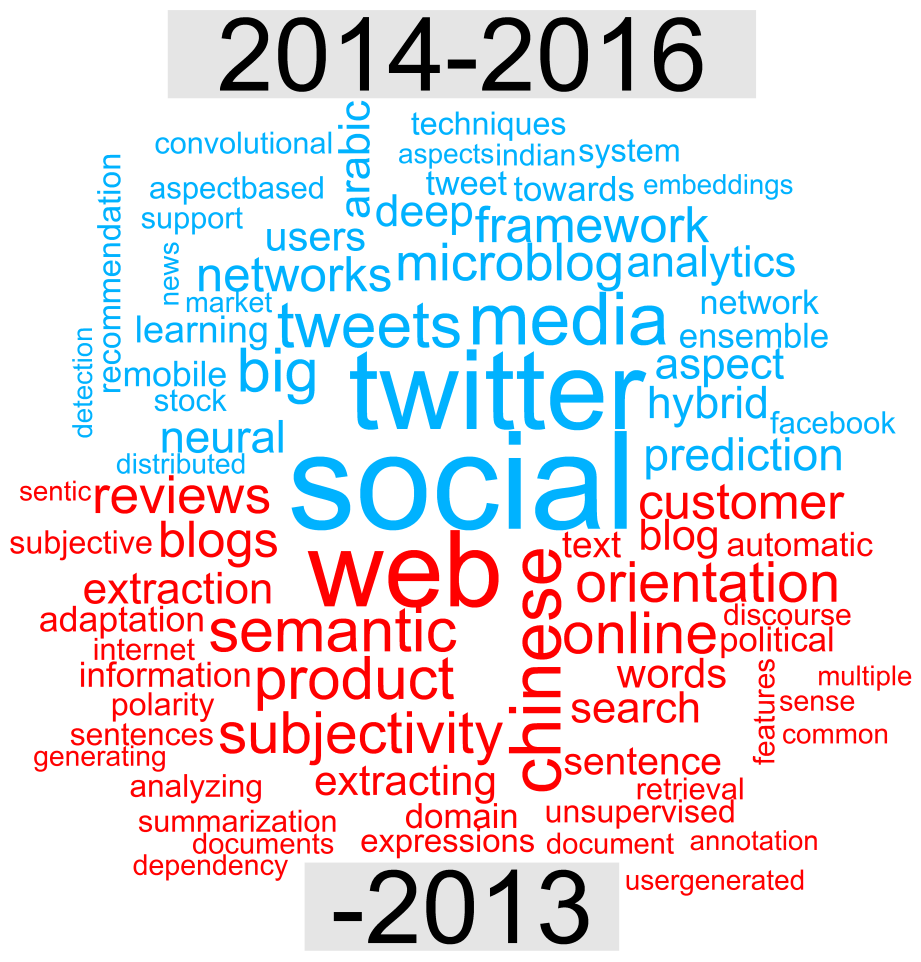

(b)

**Figure 8 a) Word cloud of all the sentiment analysis papers b) Word comparison cloud (2014-2016) papers on top and (-2013) on bottom**

### 3.5 Areas of Research - Topic modelling and human based classification of topics (RQ3)

We performed text clustering with topic modelling to study the detailed areas of research in sentiment analysis. The word clouds provide a higher level view from top down as they focus on the most frequent words. Topic modelling works from bottom up as tries to come up with a set of topics that could generate the given documents. Details of our clustering method are given in Section 2.4.

Our topic modelling computation showed that 108 was the optimal number of topics when measured with log-likelihood. We further analyzed these 108 topics with qualitative coding to form higher level groups. In comparison to our previous work [16], where we also utilized the same topic modelling approach, we noticed during the manual classification that topics in the area of sentiment analysis tended to be more incoherent in terms of the most probable words for each topic.

The quality of the topics fluctuates. There are very coherent topics like "T94 software projects usability developers infrastructure contributors developer opensource secondary maintenance" and "T80 stock price trading prices investors returns investment movement stocks forecasting". However, there are also topics that are incoherent. In comparison to our past work [16], we were less satisfied with the results of the topic modelling in here. To solve the issues of incoherence of some topics during manual qualitative coding, see Section 2.5, we coded such topics under multiple classes. Next, we present our classification from the coarsest grained groups to the fine-grained groups.

We formed three groups in the highest level, namely *Data, Data Analysis, and Goal*. Figure 9 shows out top level classification tree while Figures 10 to 14 show the details of each of each branch.

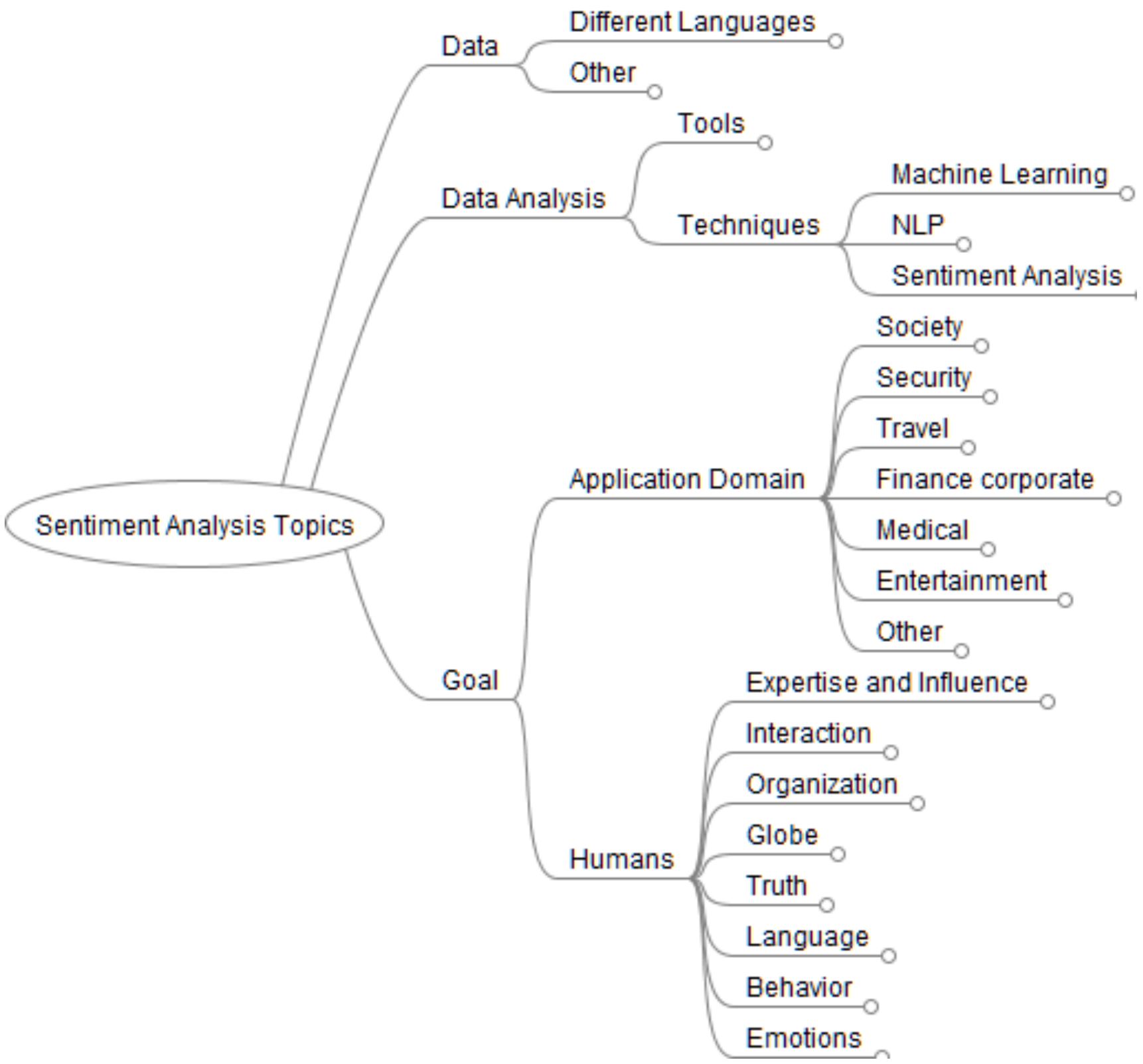

**Figure 9 Top level of our classification tree.**

### 3.5.1 Data

The *Data* area collects all topics related to the source of information that is utilized. The Data class contained 10 topics directly under it where the most probable words about data were like: feed, music, photo, image, video, microblogging, email, chat, mobile, smartphone, newspaper, and chat. We were able to distinguish just one sub class for Data named *Different Languages*.

- *Different Languages* contained topics that described the language in which the data was expressed, e.g., spanish, italian, russian, arabic, german, etc. Interestingly, English was not found to be a high probable word. The reason is likely that the English language is implicitly granted in research articles.

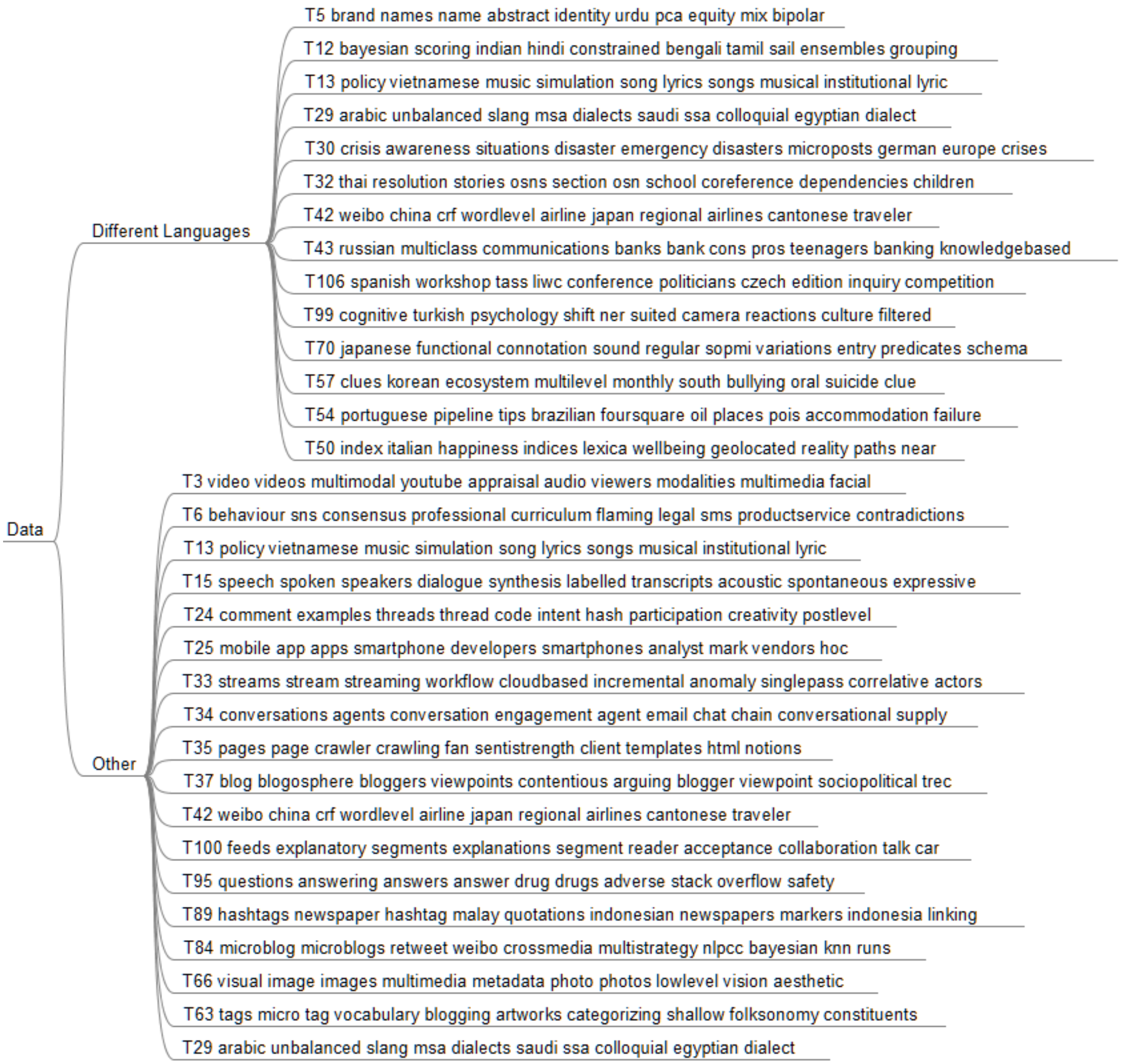

**Figure 10 Data branch of our classification tree**

### 3.5.2 Data Analysis

The *Data Analysis* class had two sub classes that contained the topics related to sentiment analysis *Tools* and *Techniques* for analyzing the data.

Tools class contained nine tool-related topics for sentiment analysis, where the most probable words were like: crawler, cloud, sentistrenght, hadoop survey, mobile. .Techniques was further divided into three sub-classes.

- *Machine Learning* had 8 topics where the most probable words were words such as: pca, bayesian, selforganizing maps, ensemble learning, knn, graphbased, fuzzy logic, crfs (conditional random fields)
- *Natural Language Processin*g (NLP) had 16 topics in which the most probable words were like: ngram, (word) embedding, bigram, stopword, parser, morphological
- *Sentiment Analysis* had 8 topics, and the most probable words were like: subjectivity, negation, sentic, dictionaries, emoticons, tone, stance, worndet

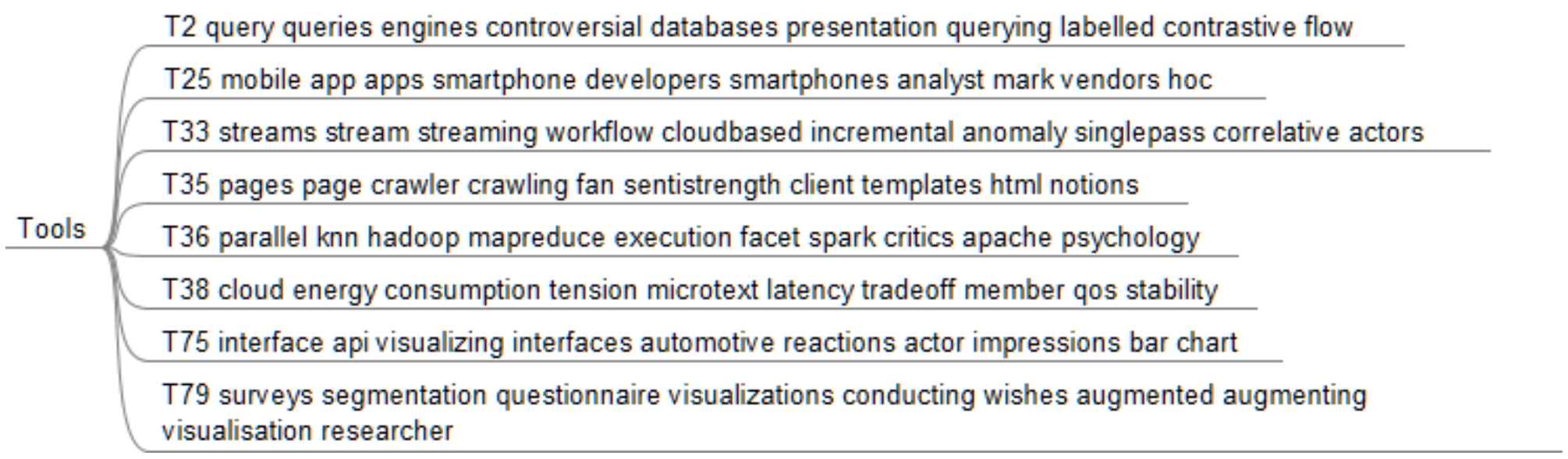

**Figure 11 Tool branch of Data Analysis branch of our classification tree**

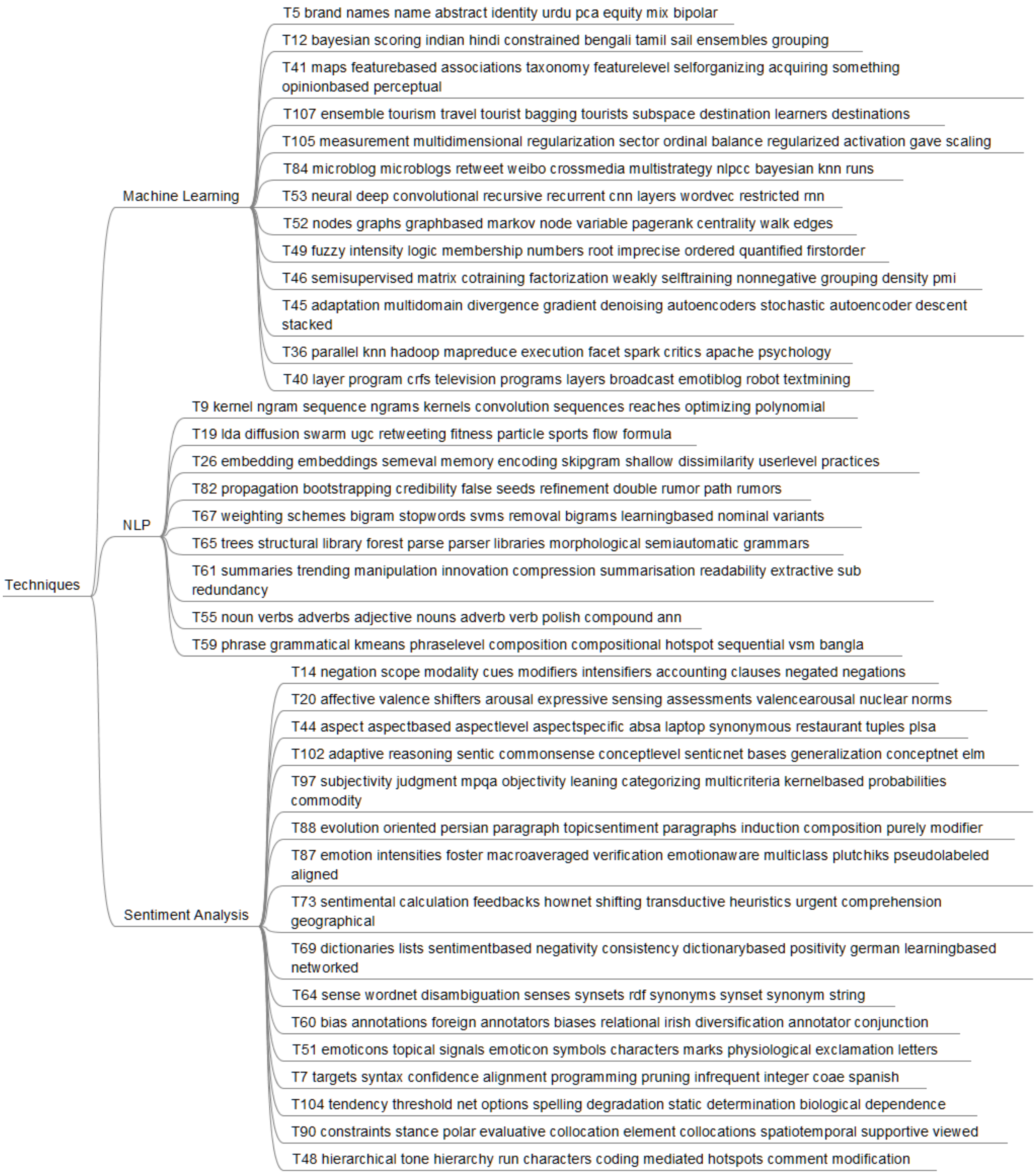

**Figure 12 Techniques branch of Data Analysis branch of our classification tree**

### 3.5.3 Goal

The goal of the paper tells what problem each paper tries to address. We identified two classes based on the types of goals: *Application Domain* oriented and *Human and Behavior* oriented.

Application domain oriented goals focused on the areas what we could call the "business" domain of sentiment analysis. It was divided further to six classes.

- *Society* had three topics where the most probable words where words like: policy, school, election, tobacco, debate, city, planning, citizens, participation.
- *Security:* had three topics where the most probable words where words like: terrorism, attacks, threats, crisis, disaster emergency, crime.
- *Travel* had four topics where the most probable words where words like: airline, travel, tourism, destination, learners, restaurant, food, hotel, tip
- *Finance and corporate* had six topics where the most probable words where words like: advertising, brand, sales, firms, banks, financial forecasting, software projects, stock price,
- *Medical* had three topics where the most probable words where words like: disease, health, patients, healthcare, drugs, suicide, depression.
- *Entertainment* had five topics where the most probable words where words like: books, imdb (international movie data base), television programs, game, player, newspaper, soccer, fan, box office.
- *Other* had six topics that each specified other application domains, such as citation analysis, education, traffic, crowdsourcing.

Human and Behavior-oriented goal focused on the areas that could be used in several application domains. Yet, we think the classes here still are research goal oriented rather than data or data analysis methods.

- *Expertise and Influence* had five topics and one sub-topic *Recommendations / Questions*. The most probable and describing words were like: expert, reputations, leader, follower, questions, ratings, recommendations.
- *Interaction* had four topics with the most probable and describing words like: discourse, arguments, audience, camera, bullying and cyberbullying.
- *Globe* had two topics with the most probable and describing words like: virtual team, temporal, spatial, geographic, international
- *Truth* had six topics with the most probable and describing words like: credibility, trust, spam, fake, irony, sarcastic, truth.
- *Language* had seven topics with the most probable and describing words like: crosslingual, writing style, domain-specific, cross-domain.
- *Behavior:* had seven topics with the most probable and describing words like: behavior, helpfulness, intentions, cognitive, reactions, determination,
- *Emotions* had seven topics with the most probable and describing words like: stress, mood, depression, valence, happiness, wellbeing,

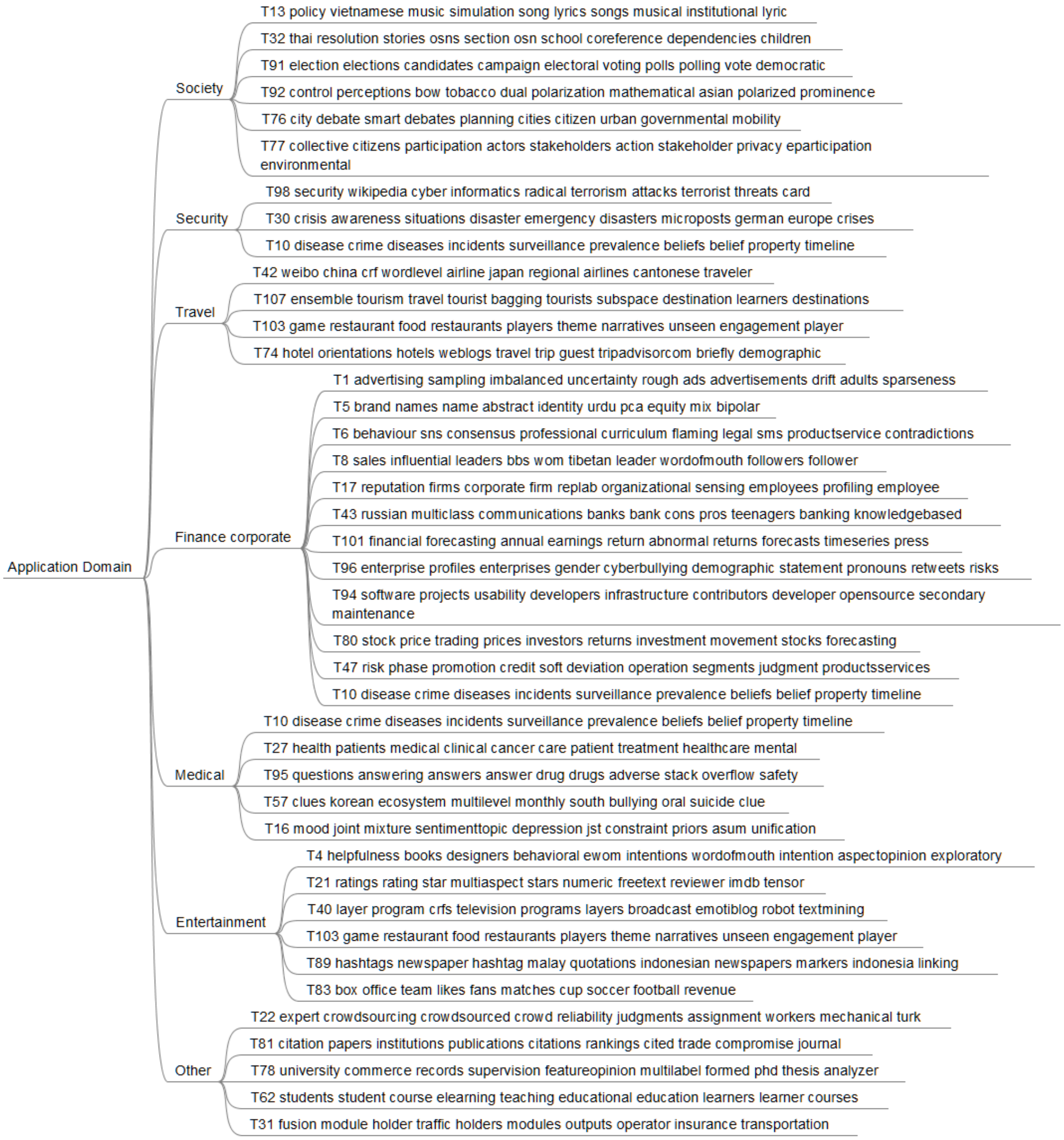

**Figure 13 Application Domain branch of Goal branch of our classification tree**

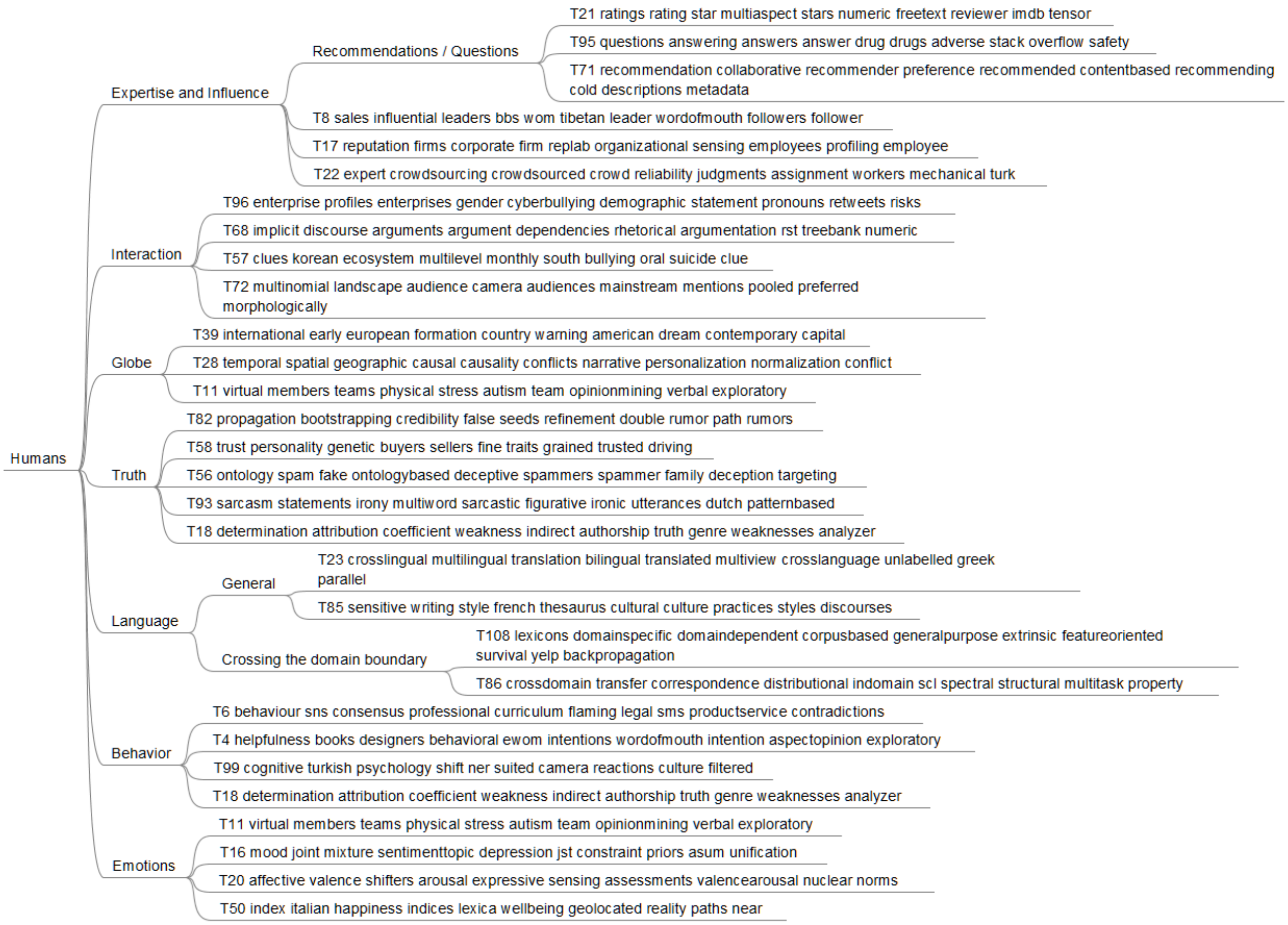

**Figure 14 Human branch of Goal branch of our classification tree**

### 3.6 Highest cited papers (RQ5)

Here we present the top-20 cited papers for sentiment analysis based on citation counts per year count. We have collected the data both using Scopus and Google Scholar, see Section 2.1 for details. For both search engines, we aimed to list top-10 papers, however, due to overlaps (marked in italics in Table 5 and Table 6), we included two extra papers from each search engine to make sure we get top-20 papers.

Table 5 and Table 6 list the top-cited papers and sub-sections 3.6.1 to 3.6.5 classify the papers into 5 groups namely: 1) Literature Reviews and Overviews, 2) Early Years – Online Reviews, 3) Twitter, 4) Tools and Lexicons, and 5) Others

**Table 5 Scopus top cited papers per year**

| Rank | Title | Year | Cites | Cites per year |
|---|---|---|---|---|
| *1* | *Opinion mining and sentiment analysis* [27] | *2008* | *2,487* | *276* |
| *2* | *Mining and summarizing customer reviews* [48] | *2004* | *1,400* | *108* |
| 3 | Lexicon-based methods for sentiment analysis [53] | 2011 | 471 | 79 |
| 4 | *Recognizing contextual polarity in phrase-level sentiment analysis [51]* | 2005 | 657 | 55 |
| 5 | Techniques and applications for sentiment analysis: The main applications and challenges of one of the hottest research areas in computer science [54] | 2013 | 203 | 51 |
| 6 | New avenues in opinion mining and sentiment analysis [55] | 2013 | 202 | 51 |
| 7 | Measuring praise and criticism: Inference of semantic orientation from association [43] | 2003 | 694 | 50 |
| 8 | Sentiment strength detection for the social web [56] | 2012 | 228 | 46 |
| 9 | Mining the peanut gallery: Opinion extraction and semantic classification of product reviews [3] | 2003 | 617 | 44 |
| *10* | *Extracting product features and opinions from reviews [57]* | 2005 | 517 | 43 |
| 11 | Estimating the helpfulness and economic impact of product reviews: Mining text and reviewer characteristics [58] | 2011 | 239 | 40 |
| 12 | Biographies, bollywood, boom-boxes and blenders: Domain adaptation for sentiment classification [59] | 2007 | 370 | 37 |

**Table 6 Google Scholar top cited per year**

| **Rank** | **Title** | **Year** | **Cites** | **Cites per year** |
|---|---|---|---|---|
| *1* | *Opinion mining and sentiment analysis* [27] | *2008* | *5,520* | *613* |
| 2 | Sentiment analysis and opinion mining [60] | 2012 | 2,027 | 405 |
| 3 | Thumbs up?: sentiment classification using machine learning techniques [41] | 2002 | 5,880 | 392 |
| *4* | *Mining and summarizing customer reviews* [48] | *2004* | *4,228* | *325* |
| 5 | Thumbs up or thumbs down?: semantic orientation applied to unsupervised classification of reviews [42] | 2002 | 4,280 | 285 |
| 6 | Twitter as a Corpus for Sentiment Analysis and Opinion Mining [61] | 2010 | 1,617 | 231 |
| 7 | Predicting elections with twitter: What 140 characters reveal about political sentiment [62] | 2010 | 1,582 | 226 |
| 8 | SENTIWORDNET: A high-coverage lexical resource for opinion mining [63] | 2007 | 2,007 | 201 |
| 9 | SentiWordNet 3.0: An Enhanced Lexical Resource for Sentiment Analysis and Opinion Mining [64] | 2010 | 1,304 | 186 |
| 10 | From tweets to polls: Linking text sentiment to public opinion time series [65] | 2010 | 1,256 | 179 |
| 11 | *Recognizing contextual polarity in phrase-level sentiment analysis [51]* | 2005 | 2,141 | 178 |
| 11 | *Extracting product features and opinions from reviews [57]* | 2007 | 1,775 | 178 |

### 3.6.1 Literature reviews and overviews

Four of the top-cited papers fall under the category of literature reviews and overviews. The paper with the highest amount of cites per year in both Scopus and Google Scholar is a literature review made by Pang and Lee [27]. The review focuses on the fundamentals and basic applications of sentiment analysis, additionally it has a list of free resources such as lexicons and data sets.

Feldman [54] introduces basic techniques and some key applications of sentiment analysis. One of the main contributions of the article is the collection of research problems the authors see the most relevant. Sentiment analysis is divided into document and sentence level analysis, while lexicon acquisition and aspect-based, aka feature based, sentiment analysis is also covered.

Cambria, Schüller, Xia and Havasi [55] give broad introductions to different techniques concerning sentiment analysis and their recent developments. Video and audio are predicted to be future data sources for sentiment analysis by the authors. Overall the paper is only seven pages long and does not go into details, thus serving better as introductory material.

One of the top cited literature reviews is a book by Bing Liu [60]. The 167 pages contain a wide array of topics, with chapters about document, sentence and aspect-based sentiment analysis. Overall the topic is approached first by introducing the research problems of sentiment analysis and then answering them with the latest knowledge available during the writing of the book.

### 3.6.2 Early years – Online reviews

Although we used citations per year count to reduce the benefit early papers gain in terms of pure citations counts, the papers from the early years that focused on online reviews still take 7 places in the top-20 cited list.

Hu and Liu [48] present a natural language based approach for providing feature-based summaries of customer reviews. The approach uses a part-of-speech tagger to divide words into lexical categories, as only the semantic orientation of adjectives is considered by the algorithm. The used algorithm is presented in detail and its implementation named "FBS" is evaluated by experiment.

Wilson, Wiebe and Hoffman [51] present phrase level sentiment analysis approach using a machine learning algorithm, which judges whether an expression is polar or neutral and the polarity of the expression. The notable contributions of this approach are the various features improving classification of sentiment polarity by taking the phrase level context into account (e.g., adverbs negating or shifting the expressed sentiment).

The article by Dave, Lawrence and Pennock [3] presents an approach to opinion mining, where the opinions of products are mined from the Web and analyzed using NLP techniques. Opinions are divided into positive and negative sentiments by the algorithm, while feature opinions and context are considered as well.

Popescu and Etzioni [57] introduce "an unsupervised information-extraction system", which improves the work of Hu and Liu [48] in the feature-based summaries of customer reviews. Main improvement to prior approaches is the use of an Internet search engine to calculate Point-wise Mutual Information (PMI) score, to evaluate if a noun can be considered a part or feature of the product.

One of the earliest works on the sentiment classification of reviews is made by Pang, Lee and Vaithyanathan [41] in 2002. Three different machine learning classifiers are used in document level sentiment analysis, particularly to analyze movie reviews and classify their overall sentiment to either negative or positive. All classifiers beat both random-choice and human-selected-unigram baselines in experimental evaluation.

Another widely cited work from 2002, concentrating on document level semantic classification, is the paper by Turney [42]. The introduced algorithm classifies the overall semantic orientation of a document based on the average semantic orientations of the phrases it consists of, using the PMI score. The approach is evaluated with reviews from different domains, i.e. automobile, bank, movie and travel reviews. Movie reviews prove to be particularly challenging for the approach, as a review of a recommendable movie can contain negative adjectives describing incidents in the movie, e.g. violence and horror.

Turney and Littman [43] evaluate two strategies for measuring semantic orientation from semantic association, i.e. statistically taking into account the context when evaluating semantic orientation. These approaches, PMI and latent semantic analysis (LSA) are tested with two different corpora, with LSA approach being more accurate in classifying semantic orientation.

### 3.6.3 Twitter

As highlighted in Figure 8 b) Twitter and social media have gained popularity in sentiment analysis in recent years. Three top-cited papers focused on working with Twitter data.

Pak and Paroubek [61] present a method for automatic collecting of a corpus from microblogs and use it to build a sentiment classifier. In this instance, the corpus is gathered from Twitter. The authors claim that the approach can be adapted to multiple languages, but in their work, it is only used with the English language.

Tumasjan et al. [62] examine around 100,000 tweets from Twitter as a predictor for election results. They find out that the proportion of mentions of parties and prominent politicians mirror the election results quite closely, however having less accuracy than opinion polls. Another interesting finding was that 40% of the messages were posted by the 4% of the users.

Data from Twitter has also been used to gauge public opinion in time series by O'Connor, Balasubramayan, Routledge and Smith [65]. While the correlation between sentiment measurement and poll data varies across different data sets, the broad trends are captured by the sentiment analysis.

#### 3.6.4 Tools and lexicons

Three different sentiment analysis tools and lexicons made it to the top-20 cited list. Such works make it easier to take sentiment analysis into to use.

Taboada et al. [53] enhance the existing analysis tool “Semantic Orientation CALculator (SO-CAL)” by supplementing rules such as intensification and negation to a pre-existing lexicon based method. This new approach is then validated, evaluated and compared with several existing sentiment analysis tools and approaches, in an experiment where various data sets are analyzed with all the approaches to draw conclusions. The authors used several data sets to evaluate the tool: product reviews, social media comments, news articles and headlines.

In 2012, Thelwall, Buckley and Paltoglou [56] evaluate a tool named “SentiStrength 2”, which is a lexicon based classifier used together with various machine learning algorithms. This tool is evaluated with a variety of data sets that feature shorts texts, e.g., YouTube comments and tweets. The tool performs well analyzing short texts found on the social web and performs fairly well context independent.

SentiWordNet is an automatically generated lexicon developed by Baccianella, Esuli and Sebastiani [63], [64] to be specifically used in sentiment analysis. The lexicon was generated using a semi-supervised learning algorithm and evaluated by comparing the generated semantic values of words with the manually annotated sets of synonyms.

#### 3.6.5 Others

Finally, we had two papers that did not fit into our existing categories but both of them addressed important topics, i.e., economic impacts and the domain dependency of sentiment analysis.

Widely cited paper by Ghose and Iperoitis [58] examined the relation of product reviews to economic outcomes and implemented a random forest based classifier to predict the impact of reviews. Additionally, they found that subjectivity of a review is associated with an increased impact of a review, while spelling mistakes decrease the economic impact of some reviews.

Work by Blitzer, Dredze and Pereira [59] concentrates on the problem of context dependent nature of sentiment expression. The authors extend a structural correspondence learning algorithm, which can adapt between domains. Additionally, the authors introduce a measure of domain similarity, to gauge the transferability of classifiers from domain to domain.

## 4 Limitations

We do realize that using our search strings in the title, abstract and key words has resulted in the inclusion of papers that only use sentiment analysis to motivate their work but we think those works are still valid as advances in, e.g., in sarcasm detection can boost advanced in sentiment analysis as well. However, we think that those papers are in the minority. This belief is based on the finding that among our top-cited literature reviews or original works, altogether a sample of 17 papers, we only found studies that were directly about sentiment analysis.

Our choice of employing the Scopus database with minor additions from Google Scholar could be considered a limitation of this study. The most widely employed academic databases are Google Scholar, Scopus, and the Web of Science [23]. There has been much discussion about which database offers the least error-prone results and the highest coverage for knowledge discovery and researchers’ assessment, e.g., [23]–[26] . As reported by Harzing and Alakangas [23], Google Scholar offers the most comprehensive coverage of the literature, but Scopus follows closely and offers quality inclusion criteria and data exporting features that Scholar does not offer. Adriaanse and Rensleigh [26] found that Scopus is the most reliable of most widely used academic databases, with least inconsistencies on author names, volume and issue number. Mongeon and Paul-Hus [25] compared Scopus and the Web of Science across the fields of natural sciences and engineering, biomedical research, the social sciences, and arts and humanity. The results of their study showed, among other results, that Scopus is capable to provide the highest field coverage with an overrepresentation of natural sciences and engineering and biomedical science, which is our case. Additionally, a major motivation for choosing Scopus was its ability to easily export high numbers of search results for further processing, a feature missing from Google Scholar and Web of Science. Thus, we believe that our choice of using Scopus augmented with some Google Scholar results, while not perfect, has been optimal for the task at hand.

It is well understood that in qualitative coding the classification is influenced by the researchers who create the classification [34]. The first author who performed the majority of the classification had educational background in computer science and has performed extensive research in software engineering focusing on affect, testing, source code maintainability, and root-cause analysis. The second author who helped with creating classification had an educational background in computer science and has performed research in software engineering and psychology with focus on affects and mixed research methods including qualitative methods. The software engineering backgrounds of both authors have likely influenced the qualitative classification result.

## 5 Discussions and Conclusions

In this article, we presented a computer-assisted literature review, using automated text clustering with manual qualitative analysis, and a bibliometric study of sentiment analysis of 6,996 papers. We investigated the history of sentiment analysis, evaluated the impact of sentiment analysis and its trends through a citation and bibliometric study, delimited the communities of sentiment analysis by finding the most popular publication venue, discovered which research topics have been investigated in sentiment analysis, and reviewed the most cited original works and literature reviews in sentiment analysis.

We found that the science of sentiment analysis and opinion mining has deep roots in the studies on public opinion analysis at the start of 20$^{th}$ century. First papers that matched our search strings were post-World War II studies that investigated the public opinion, for example towards communism, in countries recovering from the devastations of the war. Still, the topic was in hibernation until the mid-2000's when it finally emerged as an important research topic due to the need and availability of online product reviews. In 2005, only 101 papers about this topic were published while in 2015 the number was nearly 5,699. This gives us a nearly 50-fold increase in a decade making sentiment analysis undoubtedly one of the fastest growing research areas of the previous years.

We found that the citation counts have increased along with the paper counts. We found for example that the top-cited paper of sentiment analysis exceeds the citation counts of any paper published in a much mature and larger research area of software engineering. It is notable that the pool of papers used for sentiment analysis was only roughly 5,000, while our past work on software engineering had nearly 70,000 papers in the pool. Thus, sentiment analysis is also making an impact at least when measured by the number of citations.

We discovered that sentiment analysis papers are scattered to multiple publication venues and the combined number of papers in the top-15 venues only represent ca. 30% of the papers in total.

Investigation of the research topics showed that sentiment analysis had used multiple data sources related to or coming from newspapers, tweets, photos, chats for example. We found that numerous data-analysis methods had been used and we classified them to three groups namely: machine learning, natural language processing and sentiment analysis specific methods. With respect to research goals, i.e., the targets of or issues with sentiment analysis we found numerous application areas like, movies, travel, health, argumentation, interaction with audience, elections, expertise, sarcasm, spam, dialects and so on.

Investigation of changes in the research topics found that the most recent papers (2014-2016) had more focus on social media such Twitter and Facebook. Other topics which became popular in the recent years had been mobile devices, stock market, and human emotions. On the contrary, the papers published 2013 or prior to that had focused more on sentiment analysis in the context of produce reviews, product features and analysis of political situations such as elections.

We produced a comprehensive taxonomy of the research topics of sentiment analysis with text mining and qualitative coding. We believe that studies like our can be act as broad overviews of areas that are too large to be investigated with traditional literature reviews.

Shortly before the submission of the present paper, another bibliometric study on sentiment analysis was published by Piryani et al [66]. There are similarities but also many differences between our and their work. Similarities are:

- Both studies are bibliometric studies on sentiment analysis.
- Both articles show roughly over tenfold growth trends in papers per year in a decade. Paper counts from 2005 to 2015 in Piryani et al show growth trend of 1,400% (98/7) while we show growth trend of 4,139% (1,490/36)
- Both patterns analyze citation patterns but comparison cannot be made as we present citation counts and non-cited papers while they draw a map of citations

Notable differences between the papers are:

- We use Scopus supplemented with Google Scholar for top-cited papers as our data source. They use Web of Science.
- Our search identifies total of 6,996 papers from Scopus while their work finds 697 papers.
- Some differences in the search terms are that they also included terms like "affective computing" which we excluded as it often refers to sensor based measures. We have more detailed search terms about survey and text based sentiment analysis that are missing in their work e.g. "opinion analysis" and "semantic orientation"

- Both papers consider top publication venues with paper counts but their work only lists journals while our paper shows that only four of the top 15 publication venues are journals. Thus, considering only journals in their work can be seen as a limitation. Both papers show that the top journal in the area is Expert Systems With Applications, however, it is ranked as 9th overall behind many conference proceedings in our list. Decision Support Systems is identified in both, Piryani et al rank it 4th while we rank it also in 4th in journals but only 13th overall.
- Research topics are analyzed in both papers but with different approaches. They used a totally manual approach while our larger corpus forced us to use computer based clustering in the first step after which we manually formed higher level research topics. Still, the identified topics have many similarities in terms of data sources (news, reviews, blogs, twitter), data analysis (machine learning, NLP), and application domains (finance, medical, advertisement, traveling). We present our classification as a tree like structure with more details in comparison to Piryani et al. As we used computer-based methods we were to independently arrive at the same but more accurate conclusions from larger data set and with lesser effort.
- We present a more thorough review of the history of sentiment analysis
- We identify and present the top 20 cited articles of sentiment analysis.

So, what holds in the future of sentiment analysis? We assume that in the future the application areas of sentiment analysis will still increase and that the adaption with sentiment analysis techniques will become standardized part of many services and products. We see that the research methods will improve due to advances in natural language processing and machine learning. Additionally, we see a migration of currently mainly text based sentiment analysis methods towards the other affecting computing methods such as speech, gaze, and neuromarker analysis. However, we are doubtful whether sentiment analysis can achieve a similar 50-fold increase in the number of papers in the next ten years as has occurred during the past ten years (2005-2015). This is based on the fact that this would result in having over 250,000 papers on sentiment analysis published by the year 2025.

We offer closure and some future work topics to our opening quote of the idea that "The pen is mightier than the sword". As laid out in the Introduction, leaders have always been interested in public opinion, for example to increase popularity or to stump the attempts of revolution. French revolution sparked interest in public opinion and books gauging the public opinion started to appear already in early 1800's [67]. Public opinion quarterly started to appear in 1937 and only a few years earlier Nazi's had taken over in Germany. More recent events highlight future applications that could be essential in protecting democracy and civil liberties. Perhaps, sentiment analysis combined with IP-network analysis can help in detecting fake opinion post generated by a troll army, aka web-brigade [68]. Citizens of more established democracies can also benefit from sentiment analysis if it can detect opinionated or completely fake news articles that allegedly affected the presidential election of the USA [69]. Similarly, companies are influencing public opinion by generating fake news and organizations to protect their interest against scientific facts like climate change [70]. Our analysis had a branch labeled as the Truth which contains research about detecting fake and spam opinions, see Figure 14.

# 6 Acknowledgements

The first and third author have been partially supported by the Academy of Finland grant 298020. The second author has been supported by the Alexander von Humboldt (AvH) Foundation.